\documentclass[a4paper]{article}
\usepackage{arxiv}
\usepackage[utf8]{inputenc}
\usepackage{lmodern}
\usepackage[english]{babel}
\usepackage{multirow}
\usepackage{amsmath,amssymb,psfrag}
\usepackage{graphicx}
\usepackage{fullpage}
\usepackage{listings}
\usepackage{paralist}
\usepackage{appendix}
\usepackage{amsfonts}
\usepackage{comment} 
\usepackage{cancel}
\usepackage{helvet}
\usepackage{tikz,tikz-cd}
\usepackage{pgfplots,pgfplotstable}
\usepackage{lineno}
\usepackage{hyperref}
\usepackage{mathtools}
\usepackage{amsthm}
\usepackage{enumerate}
\usepackage{enumitem}
\usepackage{subfig,color}
\usepackage{epstopdf}
\usetikzlibrary{pgfplots.groupplots}
\usetikzlibrary{positioning}
\usetikzlibrary[shapes,arrows,trees]
\usetikzlibrary{matrix,decorations.pathmorphing}

\newtheoremstyle{sltheorem}
{}                % Space above
{}                % Space below
{\slshape}        % Theorem body font % (default is "\upshape")
{}                % Indent amount
{\bfseries}       % Theorem head font % (default is \mdseries)
{.}               % Punctuation after theorem head % default: no punctuation
{ }               % Space after theorem head
{}                % Theorem head spec
\theoremstyle{sltheorem}
\newtheorem{remark2}{Remark}
\newtheorem{theorem2}{Theorem}%[section]
\numberwithin{equation}{section}
\numberwithin{figure}{section}
\numberwithin{table}{section}

\modulolinenumbers[5]
\numberwithin{equation}{section}
\numberwithin{figure}{section}
\numberwithin{table}{section}
\usepackage{titlesec}
\titleformat*{\paragraph}{\itshape\bfseries\large}
\newcommand{\myref}[1]{~(\ref{#1})}

\title{Fourier Neural Networks as Function Approximators and Differential Equation Solvers}

\author{
  Marieme Ngom \\
  Mathematics and Computer Science division\\
  Argonne National Laboratory\\
  Lemont, IL 60439 \\
  \texttt{mngom@anl.gov} \\
   \And
Oana Marin \\
  Mathematics and Computer Science division\\
  Argonne National Laboratory\\
  Lemont, IL 60439 \\
  \texttt{oanam@anl.gov} \\
}

\begin{document}

\maketitle

\begin{abstract}
We present a Fourier neural network (FNN) that can be mapped directly to the Fourier decomposition. The choice of activation and loss function yields results that replicate a Fourier series expansion closely while preserving a straightforward architecture with a single hidden layer. The simplicity of this network architecture facilitates the integration with any other higher-complexity networks, at a data pre- or postprocessing stage. We validate this FNN on naturally periodic smooth functions and on piecewise continuous periodic functions. We showcase the use of this FNN for modeling or solving partial differential equations with periodic boundary conditions. The main advantages of the current approach are the validity of the solution outside the training region, interpretability of the trained model, and simplicity of use.
\end{abstract}

\section{Introduction}

The past decade has seen revolutionary advances in machine learning (ML) techniques, in particular through deep learning \cite{Geron2017}. The increasing availability of data through data-collecting initiatives and technologies has made the use of machine learning ubiquitous in image recognition and finance. One popular class of machine learning models is neural networks built to mimic the human brain. In the past 60 years a plethora of neural network architectures such as convolutional neural networks \cite{LeCun1999}, recurrent neural networks \cite{Hochreiter1997}, and autoencoders \cite{Rumelhart1986} were introduced in the literature. Depending on the task at hand, a specific architecture can prove more advantageous than another. For function approximation, for example, feedforward networks have been widely used \cite{Barron1993}, \cite{Cybenko1992}, \cite{HORNIK1989}. They are multilayered networks where the information travels from the input to the output only in the forward direction. Each layer of a feedforward network is composed of nodes linked across layers through weights and biases and activated through an activation function, for example,the ReLU, the hyperbolic tangent ($\mathrm{tanh}$), logistic, or sigmoid functions \cite{Nwankpa2018}, \cite{Glorot2010}.

The objective of this work is to build a special feedforward network that approximates periodic functions and seek periodic solutions to partial differential equations (PDEs).  We present a particular type of feedforward networks, Fourier neural networks (FNNs), which are shallow neural networks with a sinusoidal activation function. The terminology Fourier neural network is indicative of the neural network design since it mimics the Fourier decomposition as first introduced in \cite{Silvescu1999}. However, neural networks with sinusoidal activation functions were first investigated in \cite{Gallant1988}. In \cite{Liu2013}, the author presented a FNN that specializes in regression and classification tasks. The authors in \cite{Zhumekonov2019} provide a comprehensive comparative study of existing FNN architectures, and  recently another approach of constructing a Fourier operator was developed \cite{li2021fourier}. The main originality of FNNs is the nature of the activation function, which incorporates sinusoidal functions and is different from the traditional ones (ReLU, sigmoid function etc.). Many approximation theorems are available for these traditional activation functions. In \cite{Cybenko1992} and \cite{HORNIK1989}, it was proven that shallow neural networks with squashing activation functions such as the logistic or sigmoid functions could approximate any Borel measurable functions to any desired order of accuracy. In \cite{Leshno1993}, the authors showed that multilayered neural networks with nonpolynomial activation functions could approximate any function up to $O(1/N)$. In \cite{Barron1993}, the author gave universal approximation bounds for superpositions of a sigmoidal function for functions whose first moment of the magnitude distribution of the Fourier transform is bounded. With respect to FNNs,
the authors in \cite{Gallant1988} proved an approximation theorem for a squashed cosine activation function.

The current work introduces a new methodology to approximate analytic and piecewise continuous periodic functions using FNNs. The methodology is then used as groundwork to address the second objective of this paper, which is to seek periodic solutions to partial differential equations (PDEs).  Periodicity is ubiquitous in the physical  sciences (periodicity of seasons in climate science, etc.) and in the computational sciences (periodic boundary conditions in large-scale applications) and is thus a relevant computational regime. Periodic solutions of differential equations occur naturally, for example, for equations in electronics or oscillatory systems. For differential equations with periodic boundary conditions, solutions are periodic as well. Furthermore, there is a growing interest in solving differential equations using neural networks. The authors in \cite{Sirignano2018} introduced the deep Galerkin method (DGM), a mesh-free algorithm that effectively solves high-dimensional PDEs by using a deep neural network. To train the neural network, they used a loss function that incorporates the differential equations and the boundary and initial conditions. The DGM algorithm reduces the computational cost of traditional approaches, such as the finite difference method, by randomly sampling points in the computational domain instead of meshing it. In \cite{Raissi2018} and \cite{raissi2017hidden}, the authors developed a physics informed neural network (PINN) that aims at both solving and learning PDEs from data when they are not known. They showcased their network's performance by effectively solving and learning the Schrödinger, Burgers, Navier-Stokes, Korteweg-de Vries, and Kuramoto-Sivashinsky equations. A comprehensive discussion on how to solve the Poisson equation and the steady-state Navier-Stokes equation based on \cite{Raissi2018}, \cite{raissi2017hidden} and \cite{raissi2019Pinn} is provided in \cite{Dockhorn2019}. The authors in \cite{hsieh2019learning} modified existing iterative PDE solvers with a deep neural network to accelerate their convergence speed. They trained their neural network on a specific geometry, thus being able to generalize its performance to a diverse range of boundary conditions and geometries while significantly improving the speedup as compared with the performance of the original solver.

In the first part of this paper, we present the architecture of our FNN which uses the full cosine as an activation function (instead of the squashed one). Furthermore, we embed the periodicity information in the loss function, which ensures, upon convergence, that the network will approximate the Fourier series outside the training interval. We restrict ourselves to the approximation of low-frequency continuous periodic functions. As shown in \cite{Parascandolo2017}, a significant drawback in training FNNs is that the optimization can stagnate in a local minimum because of the high number of oscillations. Limiting ourselves to low-frequency modes and incorporating the periodicity in the loss function allow us to overcome that difficulty. We verify the results numerically on different functions and exploit the constructed FNN to recover low-frequency Fourier coefficients. One bottleneck of machine learning is the difficulty of preserving the ``learned'' parameters outside the training domain. Given the choice of both activation and loss function, the parameters learned by the FNN are also valid outside the training region. 

The second part of this paper focuses on using the constructed FNN to seek periodic solutions of differential equations, which is a novel use of this network design.  Here we follow \cite{raissi2019Pinn} and \cite{Sirignano2018} and incorporate the equations in the loss function to obtain a physics informed Fourier neural network (PIFNN).  We show the performance of the built PIFNN on a range of PDEs, such as the Poisson equation and the heat equation.

The paper is organized as follows. In Section\myref{sec:fnn} we construct the FNN and provide an initialization strategy for the weights and biases of the network. In Section\myref{sec:results} we present different numerical simulations that showcase the advantages of the built network. In Section\myref{sec:fnnpde} we modify the constructed network so it becomes a PIFNN that aims at seeking periodic solutions of a range of differential equations. In Section\myref{sec:conc} we summarize our conclusions and briefly discuss aveunues for future research.

\section{Fourier Neural Networks as function approximators}\label{sec:fnn}

\begin{figure}[!htb]
    \centering
    \includegraphics[width=0.45\textwidth, angle = 0 ]{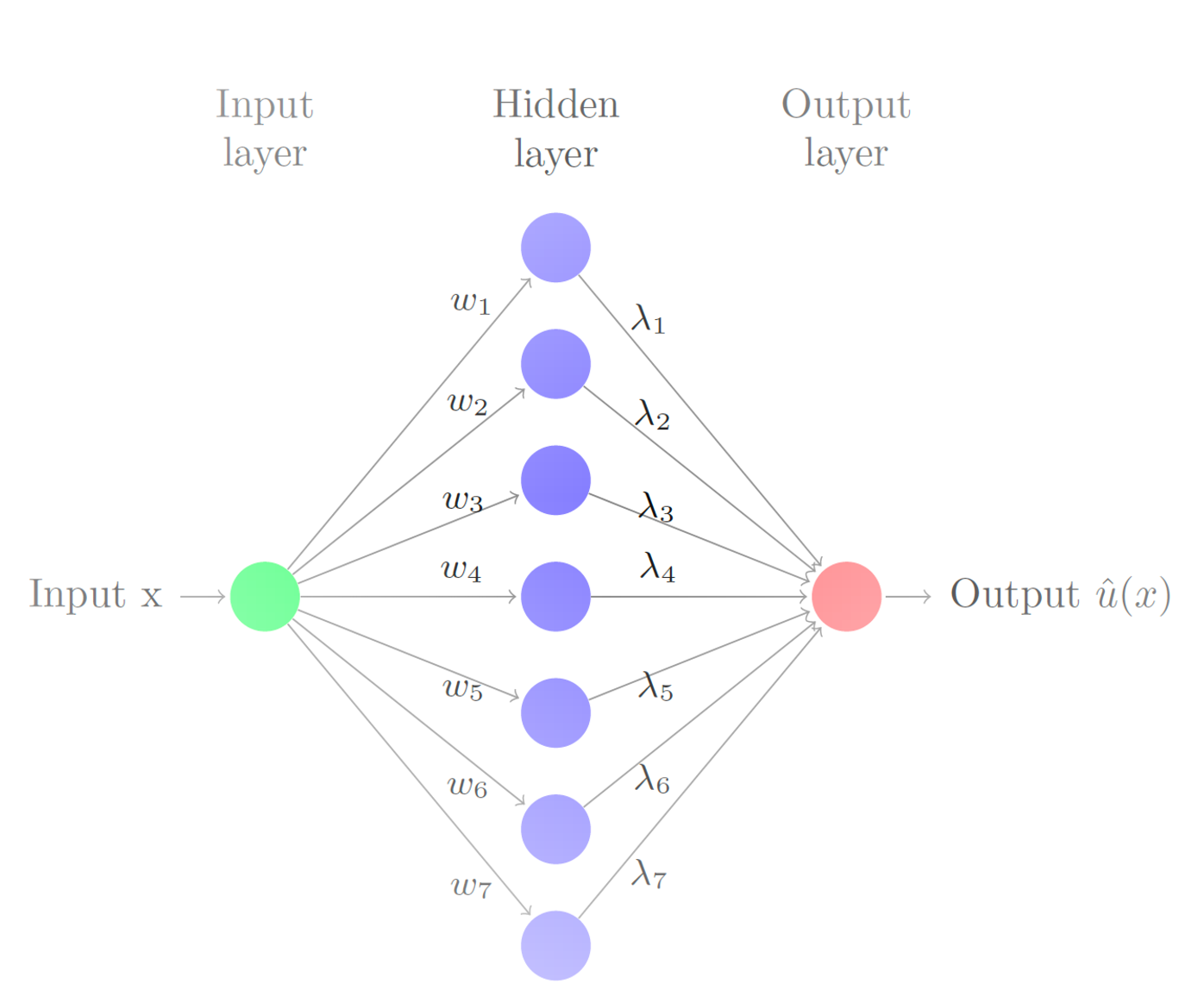}
    \caption{\;Fully connected neural network (biases not shown) with a single hidden layer and a one-dimensional input $x \in \mathbf{R}$.}
    \label{fig:NN_single}
\end{figure}

A neural network can be seen as a function approximator with a number of inputs $M$, a number of hidden layers that are composed of nodes, and a number of outputs $L$. In this work, we focus on shallow neural networks with one input, one output and $N$ nodes in the hidden layer (see Figure\myref{fig:NN_single}). The goal is to approximate real-valued periodic functions. The output $\hat{u}$ of such neural networks  can be written as
\begin{equation}\label{eq: NN1d}
  \hat{u}(x) = \phi_0 + \sum_{k = 1}^N \lambda_{k} \xi\left( w_{k}x + \phi_k \right),
\end{equation}
where $x  \in \mathbf{R}$ is the input, $w = (w_{k},\; k=1\cdots N)$ and $\lambda = (\lambda_k,\; k=1\cdots N)$ are the weights of the neural network, $\xi$ the activation function, and $\phi = (\phi_k,\; k=0\cdots N)$ its biases. 

The discrete Fourier series representation $S_N u$ of a T-periodic function $u \in L^2(\mathbf R)$ is
\begin{equation}\label{eq: fourier}
    S_{N}u(x) = \frac{a_0}{2} + \sum_{n=1}^N a_{n} \mathrm{cos}(n \omega  x) + b_{n} \mathrm{sin}(n \omega  x), 
\end{equation}
where $x \in [-\frac{T}{2}, \frac{T}{2}]$, $\omega = \frac{2\pi}{T}$, and $a_{n}$ and $b_{n}$ are the Fourier coefficients of $u$. Equation\myref{eq: NN1d} can be rewritten in the following reduced (phase shift) form:
 \begin{equation}\label{eq: fourier_shift}
     S_N u(x) = \frac{a_0}{2} + \sum_{n=1}^N c_{n} \mathrm{cos}(n \omega  x + \psi_{n}),
 \end{equation}
where for $n \geq 1$, $c_{n} = \sqrt{a_{n}^2 + b_{n}^2}$, and $\psi_n = \mathrm{arctan2}(-\frac{b_{n}}{a_{n}})$. 

Therefore, by taking the activation function to be $x \mapsto \mathrm{cos}(\omega x)$ in Equation\myref{eq: NN1d}, one obtains a formulation that resembles the Fourier decomposition. The input to hidden layer weights, $w_k$, mimics the nodes, $n$; the hidden layer to output weights, $\lambda_k$, approximates the Fourier coefficients, $c_n$; the biases of the hidden layer, $\phi_k$, correspond to the phase shifts, $\psi_k$; and the bias of the output, $\phi_0$, corresponds to half the $0^{th}$ Fourier coefficient, $a_0$. As we empirically show below, however, the equivalence is not straightforward; and one needs to train the network with a specific loss function in order to obtain satisfactory results. In the rest of this paper, we restrict ourselves to $2$-periodic functions, which means the activation function is $x \mapsto \mathrm{cos}(\pi x)$. This restriction however is made only to facilitate the design of a robust neural network and derive a reliable initialization that assures fast convergence. If the data-set is not 2-periodic it can be easily mapped to a 2-periodic set as a data processing step.

We note that Equation\myref{eq: fourier_shift} can be regarded as an interpretable neural network given its translatable relationship with Equation\myref{eq: fourier}. However no explicit translation is necessary since the node values obtained are already the modes of the Fourier series and can be directly used.

\subsection{Loss function}
The goal of this network design is to approximate the Fourier series $S_N u$ of $u$. To this end, we define the loss function as
 \begin{equation}\label{eq:lossfunction}
     L(\phi, w, \lambda) = ||\hat{u} - u ||_2^2  + \alpha_1||\lambda||^2 + \alpha_2||w||^2,
 \end{equation}
 where
 \begin{itemize}
     \item the first term $ ||\hat{u} - u ||_2^2$ ensures that the output of the neural network approximates the target function $u$, and can be regarded as a least-square estimation for obtaining Fourier coefficients \cite{Popinski1993}, 
     \item the last two terms, $\alpha_1||\lambda||^2$  and $\alpha_2||w||^2$, are regularization terms to avoid overfitting of the neural network. Choosing a $L^2$ norm for the regularization parameters causes the weights to decay asymptotically to zero, while a $L^1$ regularization might decrease them to exactly zero. We reserve the choice of the regularization norm for Section\myref{sec:results}.
 \end{itemize}
 
 However, this loss function will not provide an approximation of the Fourier series of a function unless the function is an exact combination of cosines or sines. For example, attempting to fit a piecewise continous periodic function such as 
 $$u(x) = x^2,\;\; \text{$x \in \left(-(2k+1),\;(2k+1)\right)$},\;\;k \in \mathbf{N}$$ with one neuron in the hidden layer leads to the output 
 $$\hat{u}(x) = \phi_0 - \lambda_1 \mathrm{cos}(\pi w_1 x), $$ where  $\frac{\lambda_1 \pi^2 w_1^2}{2!} \approx 1$, $\lambda_1 \approx \phi_0 \approx 1$, and $(\pi w_1)^{2l} << 1$ for $l >1$. Let us scrutinize the Taylor expansion of the chosen activation function: 
 $$
 \mathrm{cos}(\pi w_1 x) = 1 + \sum_{l=1}^{\infty} (-1)^l\frac{(\pi w_1)^{2l}x^{2l}}{(2l)!} = 1 -\frac{\pi^2 w_1^2}{2!}x^2 + o(w_1^3).
 $$
This indicates that if we seek the Fourier series of $x^2$, which is the first order term in the Taylor expansion of the activation function, the network may converge rapidly to
  $$
 x^2 \approx (1 - \mathrm{cos}(\pi w_1 x) )\frac{2!}{\pi^2 w_1^2} \approx \phi_0 - \lambda_1 \mathrm{cos}(\pi w_1 x).
 $$
This, however is not the Fourier expansion of $x^2$, which will be further studied in Section\myref{sec:results}. To alleviate such pitfalls and obtain the Fourier representation of the target function, we need to enforce the weights from the input to the hidden layer converge to integer values. This can be achieved by solving a mixed-integer optimization problem, which is out of the scope of this paper and we solve this issue here by introducing a novel family of loss functions as follows:
  \begin{equation}\label{eq:lossfunction_good}
     L(\phi, w, \lambda) = ||\hat{u}(x) - u(x) ||_2^2  + \alpha_1||\lambda||^2 + \alpha_2||w||^2 + \alpha_3\left( ||\hat{u}(x + T) - \hat{u}(x)||_2^2 \right)+ \alpha_4 \left( ||\hat{u}(x - T) - \hat{u}(x)||_2^2 \right)
 \end{equation}
This choice forces the output of the neural network to be periodic; and as we will show, this loss function preserves the periodicity of the solution outside the training interval. 
%For this loss function to force the nodes to be converge to integer values the period $T$ needs to be the minimal period, and not a multiple of it.

\subsection{Weights and biases initialization}\label{subsec:weightsini}

A proper weight initialization can significantly improve a neural network's performance in the training phase, especially when dealing with deep neural networks (DNNs). When training a DNN, one is often faced with the vanishing/exploding gradients phenomenon \cite{Geron2017}, which causes the training not to converge to a good solution. The authors in \cite{Glorot2010} proposed an initialization method using the logistic activation function that circumvents the vanishing/exploding gradients issue. In essence, they proved that the variance of each layer's outputs needs to be equal to the variances of its inputs. Furthermore, the gradient variances should be equal as the layers are traversed in the reverse direction. He et al. \cite{Heinit2015} proposed an initialization strategy for the ReLU activation function. In \cite{Kumar2017}, this was extended to nonlinear activation functions differentiable in the vicinity of the origin. 

Even though we consider a shallow neural network, initialization is essential for accelerating the training phase, and we propose a similar strategy. We initialize the biases to zero and determine appropriate weights for initialization. To this end, we revisit a few properties of the mean and variance of a random variable, denoted here by $X$ to differentiate it from its realization $x$.
Let $x$ be the input of our FNN and $\{h_k\}_{k = 1\cdots N}$ the hidden layer nodes, see Figure\myref{fig:NN_single}. For brevity, we denote by $y$ the output of the network and reserve $\hat{u}(x)$ for the Fourier series representation. Since the biases are taken to be zero, we have
$$
h_k= \mathrm{cos}(\pi w_k x ) \;\; \text{and}\;\; y = \sum_{k = 1}^N \lambda_k \mathrm{cos}(\pi w_k x). 
$$

 We assume the network is trained using $M$ samples $x$ of a random variable $X$, drawn from a uniform distribution on the period interval $\mathcal{U}(T)$, here we choose $T= 2 $ i.e $\mathcal{U}(T) = [-1, \ 1]$. A uniform distribution is a suitable choice for the input layer to assure no particular preference is given to any data. Therefore the random variable $X$ has zero mean, i.e. $\mu(X) = 0$, and variance $\sigma^2(X)=1/3$. A refresher on basic probability theory is given in Appendix\myref{appendixA}, here we define only the mean and variance as
  \begin{align}
     \mu(X) &= \int_{-\infty}^{+\infty} xf(x) dx \ ,\label{eq:mean_defs}
      \end{align}
       \begin{align}
     \sigma^2(X) &= \mu(X^2) - \left(\mu(X)\right)^2 . \label{eq:var_defs}
 \end{align}

For the hidden layer we have the set of weights; $w_k$ and for the output layer $\lambda_k$, as in Figure\myref{fig:NN_single}. Both sets are drawn from normal distributions $\mathcal N(0, m^2)$ and $\mathcal N(0, v^2)$, respectively, and we denote them by random variables $W_k$ and $\Lambda_k$, respectively i.e
$$W_k \sim \mathcal N(0, m^2) \ ,$$
and
$$\Lambda_k \sim \mathcal N(0, v^2).$$
\begin{remark2}\label{remark}
Requiring that the variances at each layer of the network are equal during the first forward pass, i.e. 
\begin{equation}
\sigma^2(H_k) = \sigma^2(Y) \ , 
\end{equation}
assures that the network initialization yields good convergence properties.
\end{remark2}
This remark allows us to determine values for $m$ and $v$, and consequently appropriate distributions for the weights $w_k$ and $v_k$, as will be shown.

\paragraph{Initialization of the hidden layer weights $\mathbf{w_k}$} 
We first compute the mean of the node $k$ of the hidden layer. To this end, we use the following theorem \cite{Ross1992}, also known as the law of the unconscious statistician (LOTUS).

\begin{theorem2}
Let X, Y be continuous random variables with a joint density function $f_{(X,Y)}$, and let h be a continuous function of two variables such that
$$\int_{\mathbf{R^2}} |h(x,y)| f_{(X,Y)}(x,y) dx dy < +\infty,\;\; \textit{then}$$
$$\mu\left(h(X,Y)\right) = \int_{\mathbf{R^2}} h(x,y) f_{(X,Y)}(x,y) dx dy  .$$
\end{theorem2}

Knowing that the joint probability distribution of the two independent random variables $W_k$ and $X$ is $$f_{(W_k,X)}(x,y) = \frac{1}{2} \cdot \frac{1}{m \sqrt{2\pi}} e^{\frac{-w_k^2}{2m^2}},$$ we obtain (using $h(w_k,x) = \mathrm{cos}(\pi w_k x)$ in the above theorem) 
\begin{equation}\label{eq:muxk}
    \mu(H_k) = \mu(\mathrm{cos}(\pi W_k X)) = \int_{-1}^{1} \int_{-\infty}^{+\infty}\frac{1}{2} \mathrm{cos}(\pi w_k x)\frac{1}{m \sqrt{2\pi}} e^{\frac{-w_k^2}{2m^2}}\; dw_k dx.
\end{equation}
Let $I_1(x) = \int_{-\infty}^{+\infty} \mathrm{cos}(\pi w_k x)\frac{1}{m \sqrt{2\pi}} e^{\frac{-w_k^2}{2m^2}}\; dw_k$. By differentiating under the integral sign we obtain
$$I_1'(x) = -\int_{-\infty}^{+\infty} \pi w_k \mathrm{sin}(\pi w_k x)\frac{1}{m \sqrt{2\pi}} e^{\frac{-w_k^2}{2m^2}}\; dw_k. $$
After integrating by parts $I_1'(x)$, we find $I_1$ satisfies the differential equation
$$I'(x) + \pi^2 m^2 x I(x) = 0, \;\; I(0) =  \int_{-\infty}^{+\infty} \frac{1}{m \sqrt{2\pi}} e^{\frac{-w_k^2}{2m^2}}\; dw_k  = \frac{m}{|m|},$$
which admits the unique solution
$$I_1(x) = \frac{m}{|m|}e^{-\frac{1}{2}\pi^2 m^2x^2} = e^{-\frac{1}{2}\pi^2 m^2x^2}. $$
We can now represent the mean as
$$ \mu(H_k) = \int_{-1}^{1}\frac{1}{2} e^{-\frac{1}{2}\pi^2 m^2x^2} dx,$$
which, after integration, gives
\begin{equation}\label{eq:muxk_last}
    \mu(H_k) = \frac{1}{m\sqrt{2\pi}}\mathrm{erf}\left(\frac{m\pi}{\sqrt{2}}\right), 
\end{equation}
where $\mathrm{erf}$ is the error function.

To compute the variance of $H_k$, we recall that $\sigma^2(H_k) = \mu(H_k^2) - (\mu(H_k))^2.$ 
Using the trigonometric identity $$h_k^2 = \mathrm{cos}^2(\pi w_k x) = \frac{1}{2} + \frac{\mathrm{cos}(2\pi w_kx)}{2},$$ we obtain
\begin{equation*}
    \mu(H_k^2) = \int_{-1}^{1}\frac{1}{2} \int_{-\infty}^{+\infty} \frac{1}{2}(1 + \mathrm{cos}(2 \pi w_k x))\frac{1}{m \sqrt{2\pi}} e^{\frac{-w_k^2}{2m^2}}\; dw_k dx.
\end{equation*}
After integration we have the mean of the random variable squared
$$\mu(H_k^2) = \frac{1}{2} + \frac{1}{4\sqrt{2\pi}}\frac{\mathrm{erf}\left(\sqrt{2}m\pi\right)}{m}.$$

The variance of $H_k$ is now readily available from Equation\myref{eq:var_defs} as
\begin{equation}\label{eq:varxk_last}
   \sigma^2(H_k) = \frac{1}{2} + \frac{1}{4\sqrt{2\pi}}\frac{\mathrm{erf}(\sqrt{2}m\pi)}{m}-\left(\frac{1}{m\sqrt{2\pi}}\mathrm{erf}\left(\frac{m\pi}{\sqrt{2}}\right) \right)^2 .
\end{equation}
%Since we require the variance of the output of the hidden layer to be equal to the variance of its input, namely, $\sigma^2(H_k) = \sigma^2(X) = 1/3$, we need to solve 
%\begin{equation}\label{eq:solm}
%   \frac{1}{3} = \frac{1}{2} + \frac{1}{4\sqrt{2\pi}}\frac{\mathrm{erf}(\sqrt{2}m\pi)}{m}-\left(\frac{1}{m\sqrt{2\pi}}\mathrm{erf}\left(\frac{m\pi}{\sqrt{2}}\right) \right)^2,
%\end{equation}
%which admits a unique solution $m \approx 0.6959$. 

\begin{figure}
  \centering \subfloat[Right hand side $s= \sigma^2(H_k) $ of Equation\myref{eq:varxk_last}.]{\includegraphics[width=0.35\textwidth]{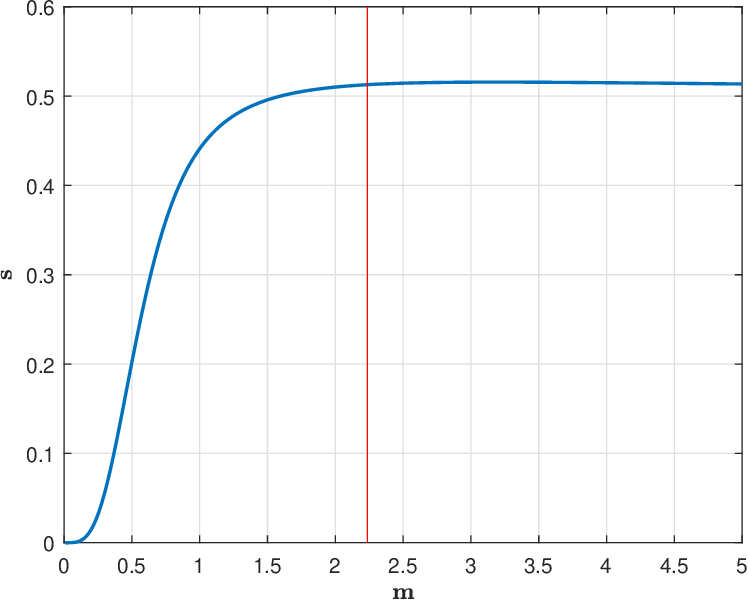}
  \label{fig:var_eq}}
  \quad \subfloat[Solution set for Equation\myref{eq:varxk_last}.]{\includegraphics[width=0.35\textwidth]{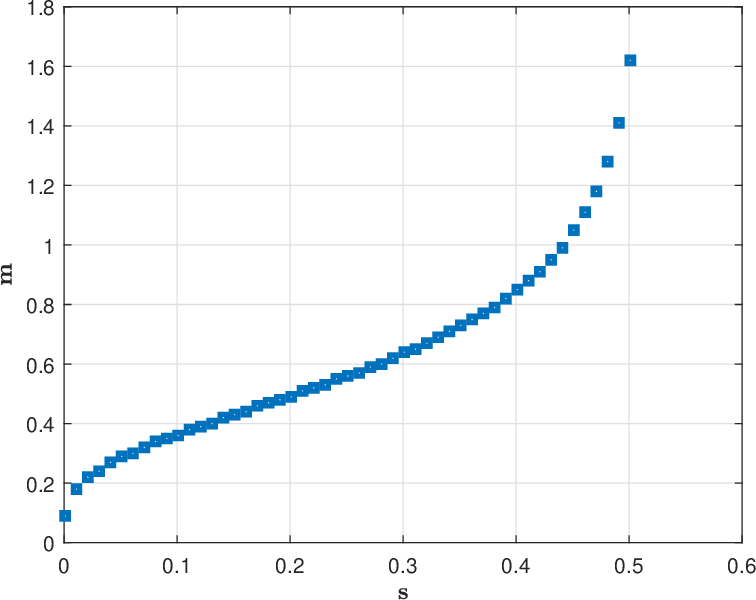}
  \label{fig:m_sol}}
  \caption{\;Analysis of the choice of hidden layer weights and their variance.}
  \label{fig:weights_init_1}
  \end{figure}

Equation\myref{eq:varxk_last} establishes a closed form expression for the connection between $ \sigma^2(H_k)$ and the variance $m$. To this end let us denote the relation between the layers of the network, stated in Remark~\ref{remark}, as $s= \sigma^2(H_k) = \sigma^2(Y)$, and refer to $s$ as the network initialization constant, 
In Figure\myref{fig:weights_init_1}  we evaluate $\sigma^2(H_k)$, as provided by Equation\myref{eq:varxk_last}, to identify the behaviour of the variances $m$ as a function of the network initialization constant $s$. We note that for values of the variance $m$ larger than approximately $\sqrt{5}$, identified via a red vertical line in Figure\myref{fig:var_eq}, the solution $s$ is asymptotically constant.  In Figure\myref{fig:m_sol} we illustrate a possible set of values for the variance $m$ given the constant $s$, and observe that the asymptotic behaviour of $s$ in Equation\myref{eq:varxk_last} implies that values $s\gtrsim 0.52$ are not feasible.

\paragraph{Initialization of the output layer weights  $\mathbf{\lambda}_k$} Since $\Lambda_k$ and $\mathrm{cos}(W_k X)$ are independent, we can write the mean and variance of the output layer $Y$
$$\mu(Y) = \sum_{k = 1}^N \mu(\Lambda_k) \mu(\mathrm{cos}(W_k X  )),$$ 
and $$\sigma^2(Y) = \sum_{k = 1}^N \sigma^2 (\Lambda_k)  \left(\sigma^2\left(\mathrm{cos}(W_k X)\right)+\mu^2\left(\mathrm{cos}(W_k X)\right) \right) +  \mu^2(\Lambda_k)\sigma^2\left(\mathrm{cos}(W_k X)\right),$$ 
    where $\sigma^2 (\Lambda_k) = v^2$ and $\mu(\Lambda_k) = 0$. This leads to
\begin{equation}\label{eq:vary}
    \sigma^2(Y) = N\left[ v^2\left[\frac{1}{2} + \frac{1}{4\sqrt{2\pi}}\frac{\mathrm{erf}\left(\sqrt{2}m\pi\right)}{m}\right] \right], 
\end{equation}
where $N$ represents the number of nodes in the hidden layer chosen at the initialization step.
Following again Remark~\ref{remark} with $s=\sigma^2(Y) = \sigma^2(H_k)$, we obtain
\begin{equation}\label{eq:varweight2}
    v^2 = \frac{s}{N\left(\frac{1}{2} + \frac{1}{4\sqrt{2\pi}}\frac{\mathrm{erf}\left(\sqrt{2}m\pi\right)}{m}\right)} .
\end{equation}
% \begin{equation}\label{eq:varweight2}
%     v^2 = \frac{F(m)}{G(m)} .
% \end{equation}

% 
% \begin{figure}
% \centering 
% \includegraphics[width=0.47\textwidth]{var_equation.eps}
% \caption{\;Right hand side $F(m)$ of Equation\myref{eq:solm}}
% \label{fig:var_eq}
% \end{figure}

 \begin{figure}
  \centering \subfloat[Right hand side $v$ of Equation~(\ref{eq:varweight2}) as function of $s=\sigma^2(Y) = \sigma^2(H_k) $.]{\includegraphics[width=0.37\textwidth]{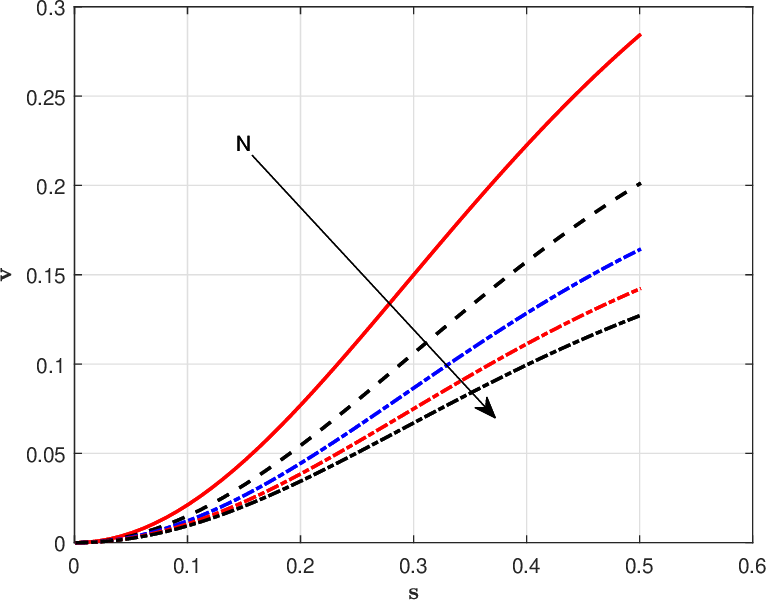}
  \label{fig:v_weights}}
  \quad \subfloat[Relationship between the variance $m$ of the hidden layer, and the output layer variance $v$ from Equation~(\ref{eq:varweight2})]{\includegraphics[width=0.37\textwidth]{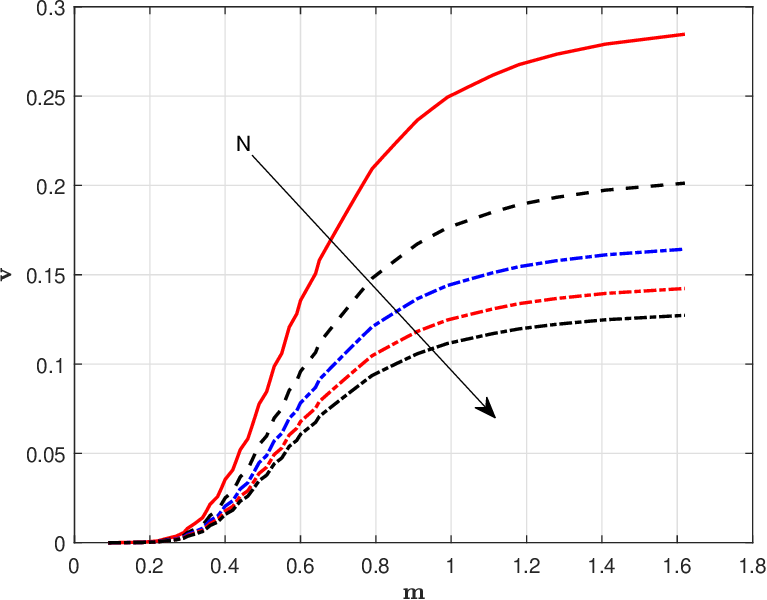}
  \label{fig:v_vs_m}}
  \caption{\; Analysis of output layer weights and their variance, browsing a set of nodes $N=[4,\ 8,\ 12,\ 16,\ 20]$ chosen for the hidden layer.}
  \label{fig:weights_init_2}
  \end{figure}

Without yet making any choice for the network initialization constant $s$, we illustrate in Figure\myref{fig:v_weights} the relationship between the variance $v$ and the constant $s$, and in Figure\myref{fig:v_vs_m} the relationship between variances $v$ and $m$. Note that in Equation~(\ref{eq:varweight2}) the variance $v$ depends on $N$, the nodes in the hidden layer, for which we select a set of values $N=[4,\ 8,\ 12,\ 16,\ 20]$. Furthermore in Figure\myref{fig:v_vs_m} we observe that too small values of $m$, up to approximately $m<0.2$ lead to almost zero variances in the output layer. This is not a desirable feature and will inform our choices. Also both in Figure\myref{fig:v_weights}-\myref{fig:v_vs_m} we note that an increasing number of nodes $N$ leads to lower values of the variances $v$, as a function of $m$. 
%
%The analysis of Equation\myref{eq:varxk_last} and Equation\myref{eq:varweight2} provides the variances to be used to initialize the weights $w_k$ and $\lambda_k$ of the network, such that Remark~\ref{remark} is satisfied. In practice we found that a uniform distribution for the input layer, i.e. $\sigma^2(X)=1/3$, and ultimately $s=1/3$ yields in Equation\myref{eq:varxk_last} the solution  $m \approx 0.6959$. 
%
%TO DO: Fix the last paragraph here
%Numerical studies performed in Section\myref{sec:results} indicated that very small values for the variance do not recover the entire set of modes of the Fourier decomposition.  As a remedy, we initialize the weights using a normal distribution $\mathcal N(0,m^2)$ with $m$ chosen to be as large as possible without $F(m)$ saturating according to Figure\myref{fig:var_eq} where $F(m)$ is the right hand side of Equation\myref{eq:varxk_last} i.e
%\begin{equation}\label{eq: Fm}
%    F(m) = \frac{1}{2} + \frac{1}{4\sqrt{2\pi}}\frac{\mathrm{erf}(\sqrt{2}m\pi)}{m}-\left(\frac{1}{m\sqrt{2\pi}}\mathrm{erf}\left(\frac{m\pi}{\sqrt{2}}\right) \right)^2.
%\end{equation}

From this analysis we conclude that good values for $m$ are in the region where $m$ is the largest possible, however not larger than approximately $2.5$, according to Figure\myref{fig:var_eq}. In practice we choose $m = \sqrt{5}$, i.e $W_k \sim \mathcal N (0, 5)$. Plugging in $m = \sqrt{5}$ in Equation\myref{eq:varxk_last} evaluates to $\sigma^2(H_k) \approx .5128$; and one further substitution in Equation\myref{eq:vary} yields $v^2= 0.9703/N$.

The choice of a fixed value for $m$ is made here to provide a general framework for the numerical studies that follow. However in particular cases, such as choices of inner nodes $N$, different periodic intervals etc., better convergence properties may be achieved for different feasible values of $m$. 

% Since these values are small, however, imposing this type of initialization on the hidden layer weights will cause the FNN to capture fewer Fourier modes than we are seeking. Therefore, we initialize these weights using a normal distribution $\mathcal N(0,5)$, which means $m^2 = 5$ and justify this choice as follows. We denote $F(m)$ the right hand side of Equation\myref{eq:varxk_last} i.e
% \begin{equation}\label{eq: Fm}
%     F(m) = \frac{1}{2} + \frac{1}{4\sqrt{2\pi}}\frac{\mathrm{erf}(\sqrt{2}m\pi)}{m}-\left(\frac{1}{m\sqrt{2\pi}}\mathrm{erf}\left(\frac{m\pi}{\sqrt{2}}\right) \right)^2,
% \end{equation}

% We observe in Figure \myref{fig:var_eq} that the function $F(m)$ has an asymptote for 
% This value was picked by trial and error and, as shown in Section\myref{sec:results}, allows us to recover the first five Fourier coefficients of a periodic function. Plugging in $m$ in Equation\myref{eq:varxk_last} gives $\sigma^2(H_k) \approx .5128$; and setting $\sigma^2(Y) = \sigma^2(H_k)$, we get from Equation\myref{eq:vary} 
% \begin{equation*}
%     v^2 = \frac{0.5128}{N\left(\frac{1}{2} + \frac{1}{4\sqrt{2\pi}}\frac{\mathrm{erf}\left(\sqrt{2}m\pi\right)}{m}\right)} .
% \end{equation*}
% Numerical studies performed in Section\myref{sec:results} indicated that very small values for the variance do not recover the entire set of modes of the Fourier decomposition.  As a remedy, we initialize the weights using a normal distribution $\mathcal N(0,m^2)$ with $m$ chosen to be as large as possible according to Figure\myref{fig:var_eq}. 

\section{Numerical studies}\label{sec:results}
To assess the effectiveness of the FNN in approximating periodic functions, we perform numerical studies on both periodic smooth functions and functions periodically extensible via continuity. First, we approximate analytic periodic functions, given as linear combinations of sines and cosines. High accuracy is expected on such functions since we have direct access to the Fourier modes. Second, we recover low-frequency Fourier coefficients of piecewise continuous periodic functions, such as $x  \mapsto |x|$ and $x  \mapsto x^2$. These functions are of particular interest since they are prevalent in fields such as acoustics and electronics \cite{Benson2013}. As expected, non-smooth periodic functions place an additional strain on the training of the FNN, either in terms of iterations to convergence or in meeting the tolerance threshold. However, the results are qualitatively satisfactory and free of numerical artifacts.
\begin{remark2} We note a limitation of our model, which is that the loss function in Equation\myref{eq:lossfunction_good} requires the minimal period $T$ of the data. This can be either identified from the dataset or, for an analytic signal as used here for validation, directly from the mathematical expression.
\end{remark2}
% We note two major limitations/features of the current FNN and the means to overcome them
% \begin{itemize}
% \item  %Note that using a multiple of the minimal period $nT$ does not yield integer values for the mode coefficients.
% \end{itemize}

\subsection{Analytic Periodic Functions}
%  It is important to note that the FNN is relevant for Fourier expansions with more than one mode, since only one mode yields a one node network easy to solve via traditional techniques.
The design of the FNN lends itself naturally to simple linear combinations of Fourier modes. We first assess a function of period $T=2$ given by
\begin{equation}\label{eq: per_onemode}
  f(x) = \mathrm{cos}(\pi x) + \mathrm{sin}(\pi x)\; .   
\end{equation}

 \begin{table}
  \begin{center}
\begin{tabular}{ |c|c|c|c|c| } 
  \hline
  \multicolumn{3}{|c|}{Number of iterations} & $189$  \\
\hline
  \multicolumn{3}{|c|}{Loss Function (upon convergence)} & $2e-4$  \\
\hline
\hline
$w_k$ & $\phi_k$ & $\lambda_k$& $\phi_0$ \\
\hline
$1.00000000$ & $-0.78539816 \approx -\pi/4$ &$1.41421354 \approx \sqrt{2}$& \\ 
$-5.96856597e-7$&$-1.00911547$ & $-2.60499122e-5$& $-1.81893539e-5$ \\ 
$1.22755482e-6$& $1.87773726$ & $-4.76966579e-5$& \\ 
$4.59348246e-8$& $-6.38405893 $ & $1.77429347e-5$& \\ 
\hline
\end{tabular}
\vspace{3mm}
\caption{\;Number of iterations, value of the loss function at convergence, and optimal weights and biases of the FNN to approximate $ f(x) = \mathrm{cos}(\pi x) + \mathrm{sin}(\pi x)$, $k = 1\ldots4$ with a $L^2$ regularization.}\label{tab:tabcossinL2}
\end{center}
\end{table}

 \begin{table}
  \begin{center}
\begin{tabular}{ |c|c|c|c|c| } 
  \hline
  \multicolumn{3}{|c|}{Number of iterations} & $87$  \\
\hline
  \multicolumn{3}{|c|}{Loss Function (upon convergence)} & $3e-4$  \\
\hline
\hline
$w_k$ & $\phi_k$ & $\lambda_k$& $\phi_0$ \\
\hline
$1.00000033$ & $-0.78540193 \approx -\pi/4$ &$1.41421847 \approx \sqrt{2}$& \\ 
$1.4969094$&$-0.71483098$ & $-4.32639025e-6$& $1.58219741e-5$ \\ 
$0.05095418$& $ 0.96613253$ & $6.73308554e-5$& \\ 
$0.14126515$& $-5.84166681 $ & $-5.13186621e-5$& \\ 
\hline
\end{tabular}
\vspace{3mm}
\caption{\;Number of iterations, value of the loss function at convergence, and optimal weights and biases of the FNN to approximate $ f(x) = \mathrm{cos}(\pi x) + \mathrm{sin}(\pi x)$, $k = 1\ldots4$ with a $L^1$ regularization. }\label{tab:tabcossinL1}
\end{center}
\end{table}

Although this function has only one mode, we build the network with four nodes in the hidden layer and seek to observe whether the redundant nodes converge to zero. To investigate the impact of different regularization terms in the loss function, we consider both $L^2$ and $L^1$ norms in Equation\myref{eq:lossfunction_good}.
We report the network converged parameters in Tables\myref{tab:tabcossinL2}--\myref{tab:tabcossinL1}, along with the number of iterations and the minimized loss function. Analytical work indicates that the expected output of the FNN is
$$\hat{u}(x) \approx \sqrt{2}\mathrm{cos}(\pi x - \frac{\pi}{4}),$$ 
which is the reduced form of  $f(x)= \mathrm{cos}(\pi x) + \mathrm{sin}(\pi x)$. The optimization converged to approximately $2e-4$ in $189$ iterations for the $L^2$ regularization, as  illustrated in Table\myref{tab:tabcossinL2}. The convergence rate is faster for the $L^1$ regularization, 
providing in $87$ iterations a loss function of magnitude order $3e-4$, as in Table\myref{tab:tabcossinL1}. Although the $L^1$ regularization displays better convergence properties, the weights themselves do not converge to exactly zero as expected (see the weights corresponding to nodes $2-4$ in Table\myref{tab:tabcossinL1}). 
 \begin{figure}
    \centering
    \includegraphics[width=0.45\textwidth]{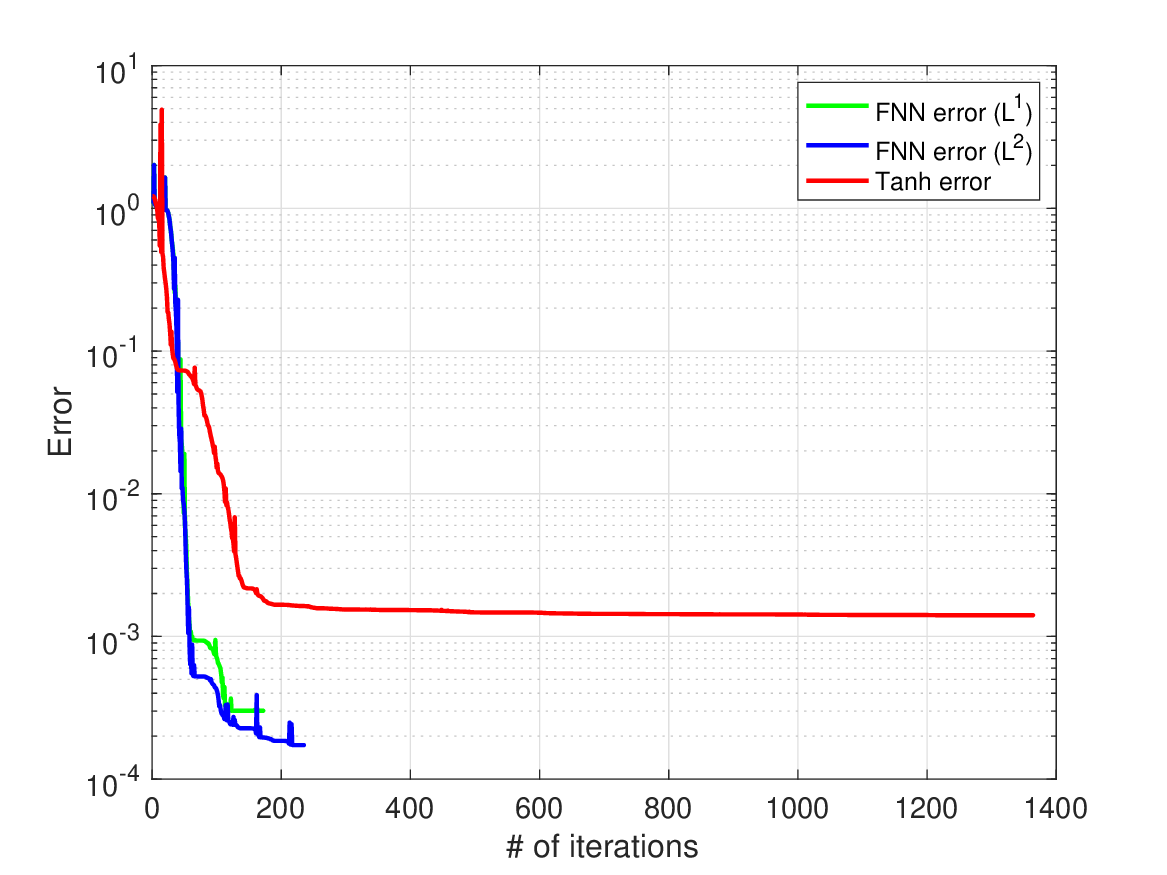}
    \caption{\;Comparison in terms of the values of the loss function throughout the optimization when using an FNN and when using a neural network with a $\mathrm{tanh}$ activation.}
    \label{fig:fourvsNN_iter}
\end{figure}

   \begin{figure}
    \centering
    \subfloat[$L^2$ regularization]{\includegraphics[width=0.45\textwidth]{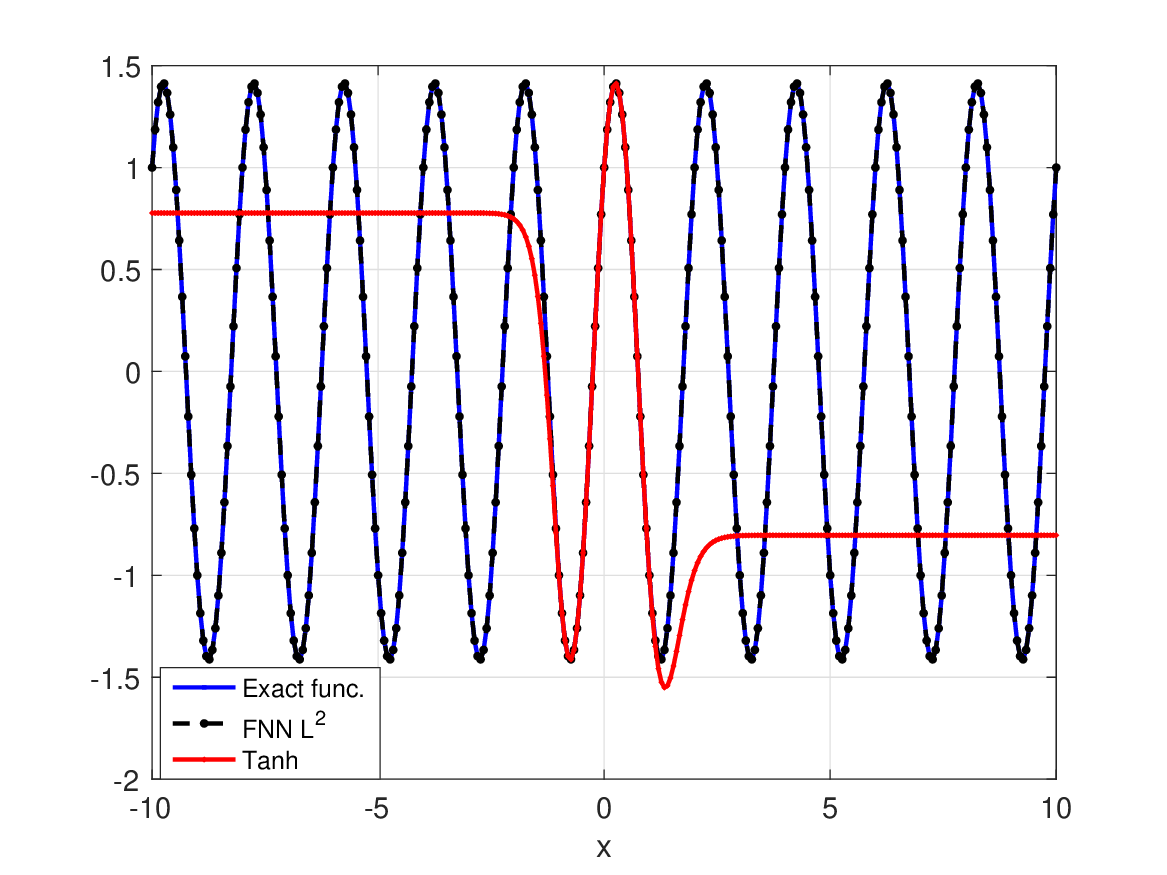}\label{fig:fourvsNN_outside_L2}}
     \subfloat[$L^1$ regularization]{\includegraphics[width=0.45\textwidth]{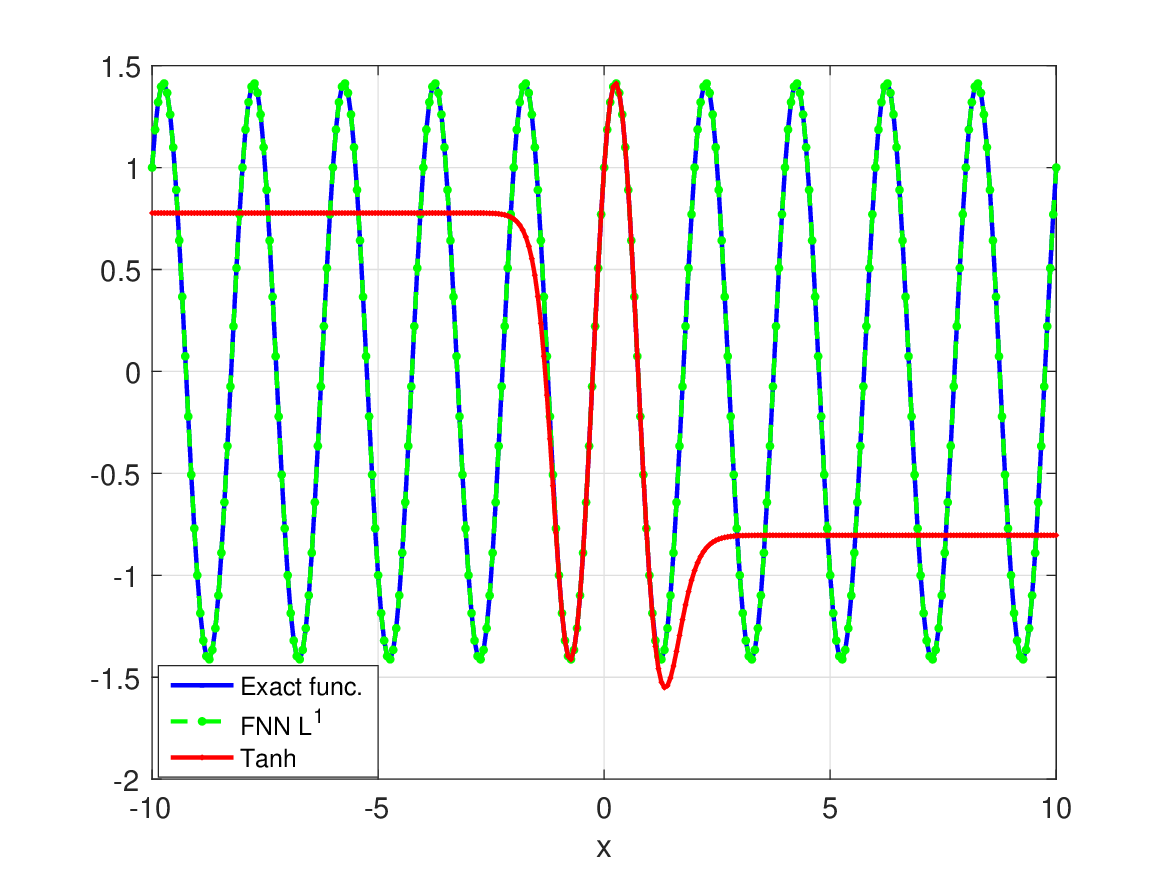}\label{fig:fourvsNN_outside_L1}}
    \caption{\;Comparison between $f(x) = \mathrm{cos}(\pi x) + \mathrm{sin}(\pi x)$ and the output of the FNN  outside of the training domain for different regularization norms.}
    \label{fig:fourvsNN_outside}
\end{figure}

In Figure\myref{fig:fourvsNN_iter} we compare the performance of the newly constructed FNN with a neural network with a $\mathrm{tanh}$ activation function and a Glorot initialization, as in \cite{Glorot2010}. The $\mathrm{tanh}$ activation function yields a slow convergence of $1173$ iterations to a loss function of $1e-3$, considerably inferior to the performance of the FNN in terms of both efficiency and accuracy. Beyond performance the FNN has the added benefit of producing an approximation valid outside the training region. In Figure\myref{fig:fourvsNN_outside} we extend the testing domain and note that periodicity is preserved for the FNN model, while the $\mathrm{tanh}$ network loses extensibility. The  inability to preserve the sought function outside of the training region is a typical shortcoming of neural network methods for function approximation. The FNN model devised here is able to overcome this  issue by imposing periodicity in the loss function, defined in Equation\myref{eq:lossfunction_good}, as well as by using a periodic activation function. We note that we did not include the periodicity requirement in the loss function when training the traditional neural network. The results were worse in that this required more nodes in the hidden layer; convergence was reached more slowly; and the error was similar. Furthermore, the traditional network did not preserve the sought function properties outside of $[-1\pm T, 1\pm T]$ when the periodicity requirement was added.

We now consider a function with more modes and also higher amplitudes:
\begin{equation}\label{eq: per_moremodes}
  g(x) = 8 \mathrm{cos}(4\pi x) + \mathrm{sin}(2\pi x) + \mathrm{sin}(\pi x).
\end{equation}

% A higher amplitude places high demands on grid resolution, and the Nyquist\footnote{insert ref \textcolor{blue}{mng: A ref about these frequencies?}} frequency is used to assure a function is properly represented. Here we use a mesh-free strategy for training the FNN and  
We show in Tables\myref{tab:tabpercompL2} --\myref{tab:tabpercompL1} the value of the loss function at  convergence, the number of iterations, and the values of the converged weights. The optimization converged to approximately $9e-4$ after $195$ iterations using a $L^2$ regularization and to approximately $1e-3$ after $169$ iterations using the $L^1$ norm.  We note that the biases are approximations of odd multiples of $\pi/2$ for the sine part of the function $g$ and approximately $0$ for its cosine component. 
 \begin{table}
  \begin{center}
  \begin{tabular}{ |c|c|c|c|c| } 
  \hline
  \multicolumn{3}{|c|}{Number of iterations} & $195$  \\
\hline
  \multicolumn{3}{|c|}{Loss Function (upon convergence)} & $9e-4$  \\
\hline
\hline
$w_k$ & $\phi_k$ & $\lambda_k$& $\phi_0$ \\
\hline
$1.00000000$ & $1.57079627 \approx \pi/2$ &$-9.99999993e-1$& \\ 
$5.40604942e-6$&$1.01888743e1$ & $3.34351764e-6$& $2.41330937e-6$ \\ 
$4.00000000$& $-9.14402304e-10$ & $7.99999983$& \\ 
$ -2.00000000$& $4.71238899 \approx 3\pi/2$ & $-9.99999984e-1$& \\ 
\hline
\end{tabular}
\vspace{3mm}
\caption{\;Number of iterations, value of the loss function at convergence, and optimal weights and biases of the FNN to approximate $ g(x) = 8 \mathrm{cos}(4\pi x) + \mathrm{sin}(2\pi x) + \mathrm{sin}(\pi x)$ with $L^2$ regularization, $k = 1\ldots4$.}\label{tab:tabpercompL2}
\end{center}
\end{table}

 \begin{table}[!h]
  \begin{center}
  \begin{tabular}{ |c|c|c|c|c| } 
    \hline
\multicolumn{3}{|c|}{Number of iterations}& $169$  \\
\hline
  \multicolumn{3}{|c|}{Loss Function (upon convergence)} & $1e-3$  \\
\hline
\hline
$w_k$ & $\phi_k$ & $\lambda_k$& $\phi_0$ \\
\hline
$0.99999306$ & $1.57072838 \approx \pi/2$ &$-1.000141261$& \\ 
$1.99999672$&$7.85406634 \approx 5\pi/2$ & $-9.99825909e-1$& $3.82384745e-5$ \\ 
$3.99999998$& $-1.14376588e-5$ & $7.99989569$& \\ 
$ -0.04032155$& $-2.12089270e+1$ & $1.69927207e-5$& \\ 
\hline
\end{tabular}
\vspace{3mm}
\caption{\;Number of iterations, value of the loss function at convergence, and optimal weights and biases of the FNN to approximate $ g(x) = 8 \mathrm{cos}(4\pi x) + \mathrm{sin}(2\pi x) + \mathrm{sin}(\pi x)$ with $L^1$ regularization, $k = 1\ldots4$.}\label{tab:tabpercompL1}
\end{center}
\end{table}

 \begin{figure}
    \centering
    \subfloat[$L^2$ regularization]{\includegraphics[width=0.45\textwidth]{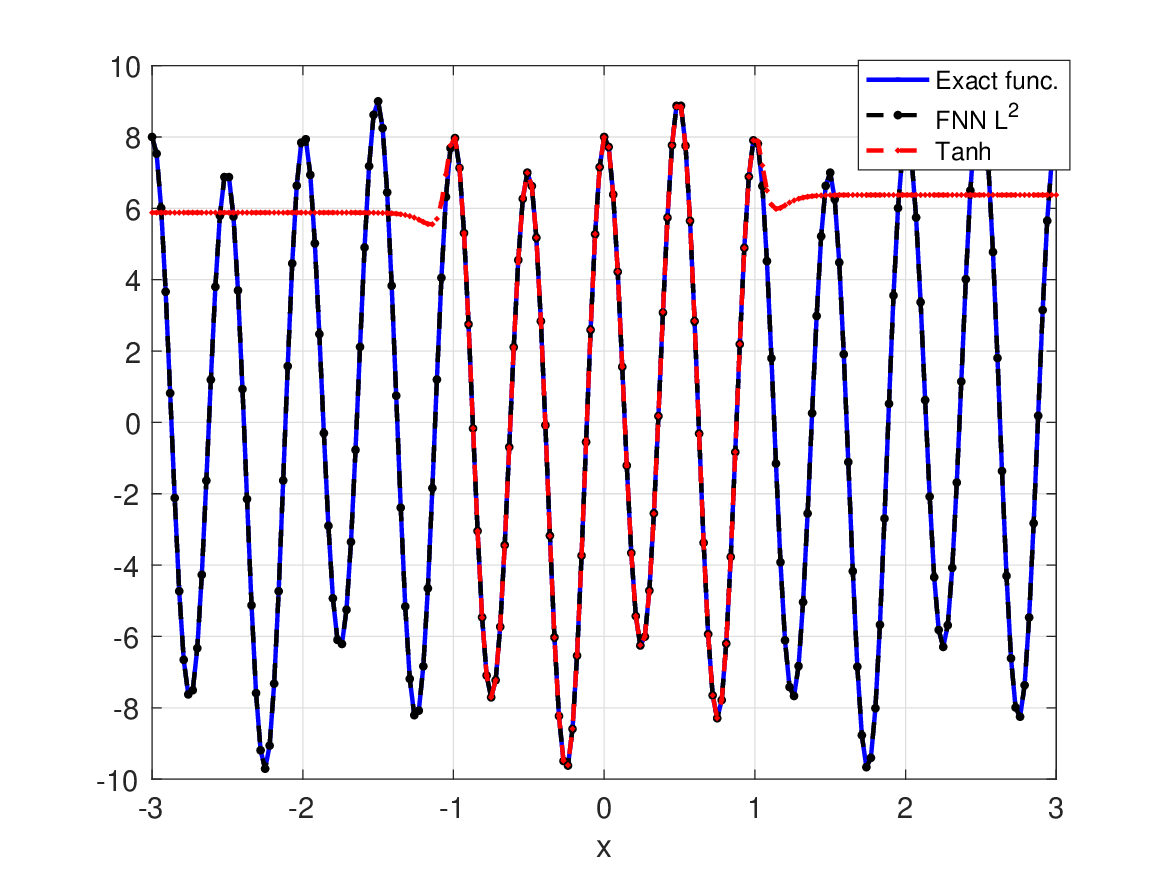}\label{fig:fourvsNNcompPer_L2}}
     \subfloat[$L^1$ regularization]{\includegraphics[width=0.45\textwidth]{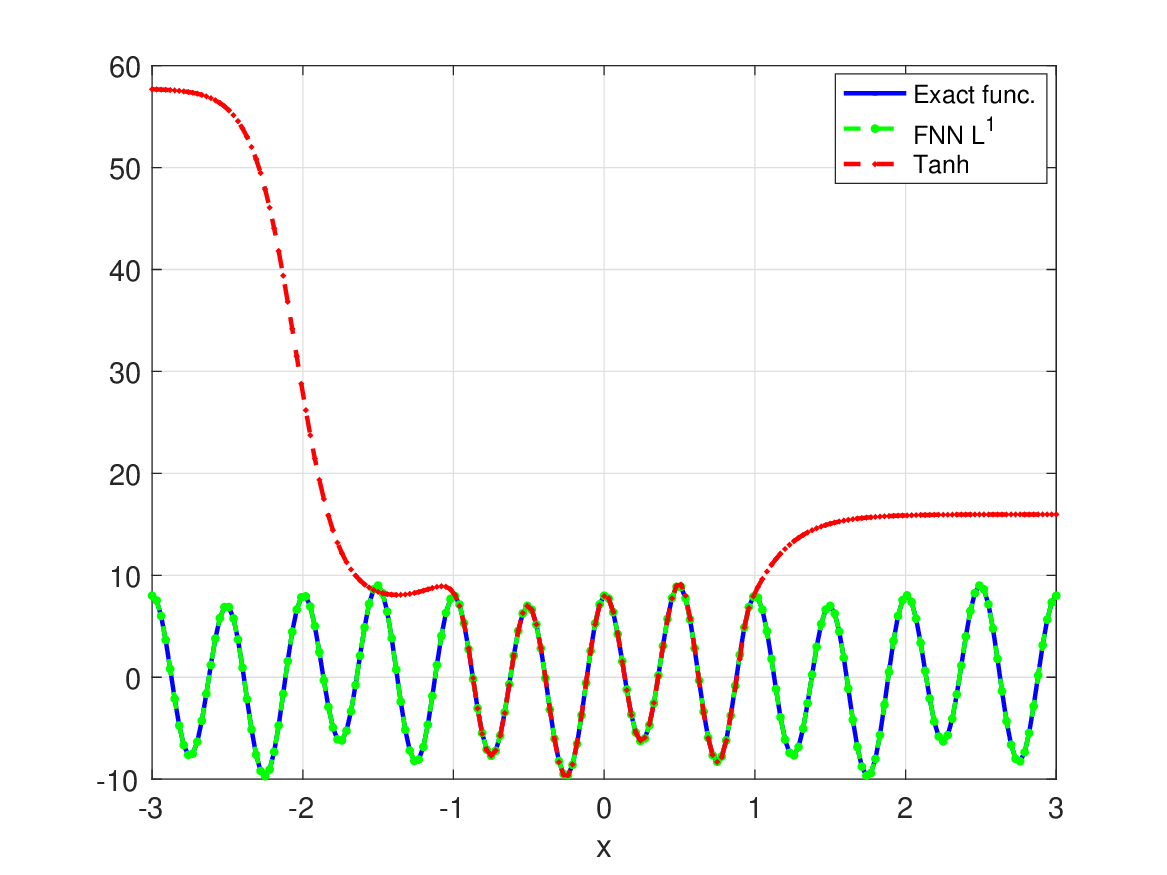}\label{fig:fourvsNNcompPer_L1}}
    \caption{\;Comparison between $ g(x) = 8 \mathrm{cos}(4\pi x) + \mathrm{sin}(2\pi x) + \mathrm{sin}(\pi x)$ and the output of the FNN for different regularization norms.}
    \label{fig:fourvsNNcompPer}
\end{figure}

In Figure\myref{fig:fourvsNNcompPer} we illustrate the comparison between the exact function $g$, the approximations provided by the FNN and by the $\mathrm{tanh}$ network. The actual error between the function and the FNN approximation has an order of magnitude $1e-8$ for the $L^2$ regularization and $1e-5$ for the $L^1$ one. Performance, as well the lack of convergence of redundant nodes to zero using $L^1$ regularization, motivates us to use $L^2$ regularization in the remainder of the paper. Comparing the FNN approximation with its counterpart obtained by using a $\mathrm{tanh}$ neural network, we note again a poor convergence for the $\mathrm{tanh}$ activation function. The local minimum for a 4-node hidden layer using the $\mathrm{tanh}$ activation has an order of magnitude 1, implying lack of convergence, while for $13$ nodes in the hidden layer convergence is attained at a value of $1e-2$ for the loss function. The FNN approximates the sought function outside the training region while the $\mathrm{tanh}$ does not. We note that we sampled the same number of points on the training interval for the function $g$ in Equation\myref{eq: per_moremodes} as for the function $f$ defined in Equation\myref{eq: per_onemode}, which has only one mode.

\subsection{Piecewise continuous periodic functions}  

We now seek to approximate nonperiodic functions, which are periodically extended through continuity. The parabola
\begin{equation}\label{eq:xsquare}
  f(x) = x^2, \;\; \text{$x \in \left(-(2k+1), (2k+1)\right)$}\;\;, k \in \mathbf{N}  
\end{equation}
has the Fourier series
$$
S_N f (x)= \frac{1}{3} + \sum_{n=1}^{N} 4 \frac{(-1)^n}{\pi^2 n^2} \mathrm{cos}(\pi nx).
$$

\begin{figure}
    \centering
    \subfloat[$f(x) = x^2$]{\includegraphics[width=.45\textwidth]{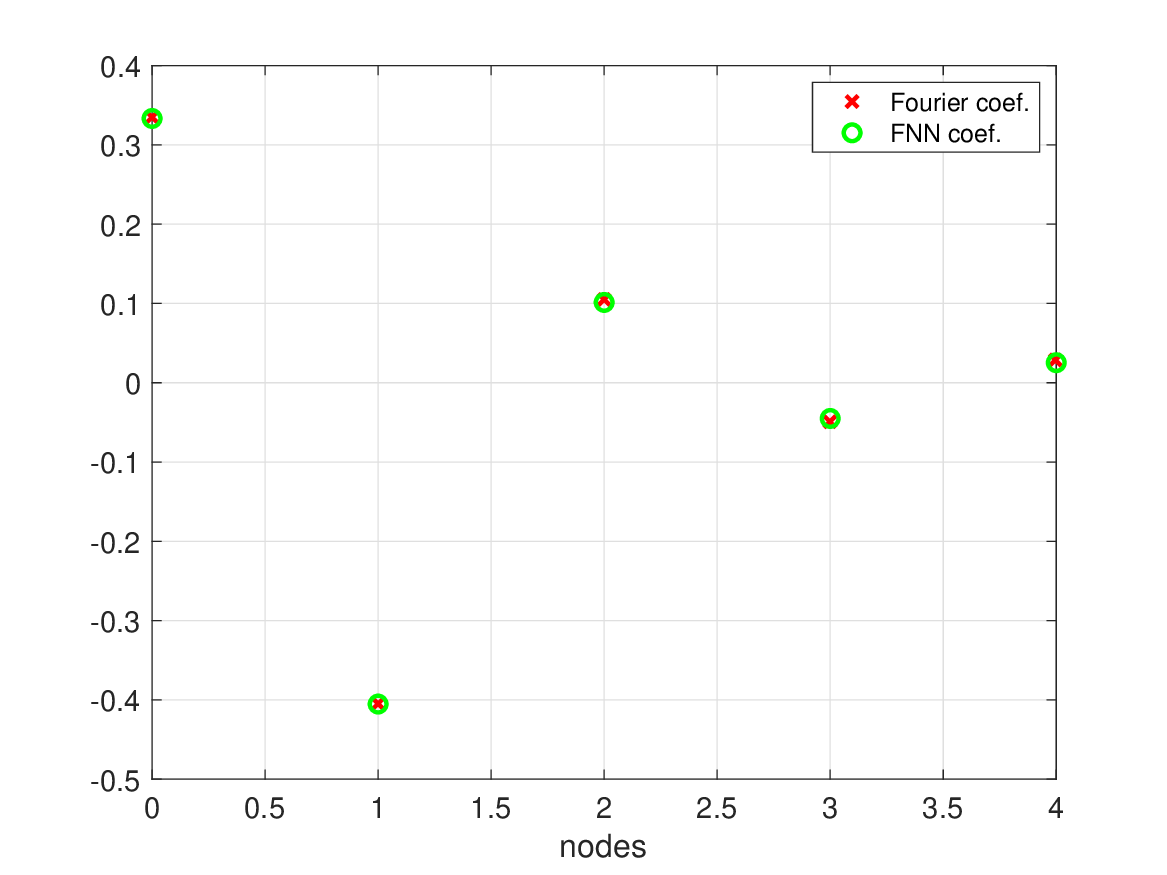}\label{fig:Fnnx2}}
     \subfloat[$g(x) = |x|$]{\includegraphics[width=.45\textwidth]{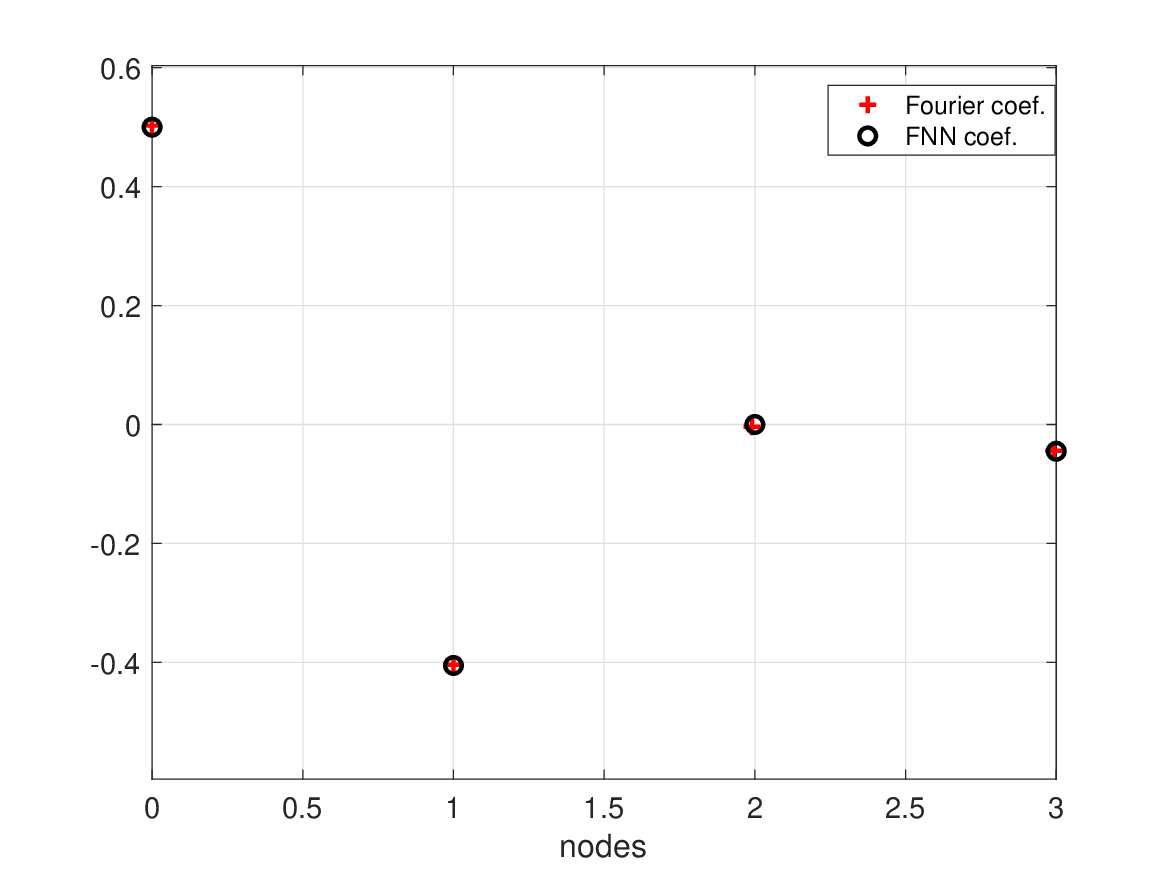}\label{fig:absx}}
    \caption{\;Comparison between the FNN coefficients and the Fourier coefficients corresponding to the nodes captured by the neural network for the functions studied.}
    \label{fig:Fnn_fg}
\end{figure}

Using $4$ nodes in the hidden layer, the FNN captures the first five modes of the Fourier decomposition (if we count the $0$th mode). We show in Table\myref{tab:tabfnnx2} the values obtained for the weights and the biases. We denote by FNN coefficients the hidden layer to output weights and the bias of the output layer. We note that they are approximations of the Fourier coefficients with an order of magnitude of $1e-3$. Figure\myref{fig:Fnnx2} shows an error bar plot for the exact Fourier coefficients against the error from the FNN coefficients and provides a qualitative assessment of the approximation.

\begin{table}
  \begin{center}
  \begin{tabular}{ |c|c|c|c|c| } 
    \hline
\multicolumn{3}{|c|}{Number of iterations}& $130$  \\
\hline
  \multicolumn{3}{|c|}{Loss Function at convergence} & $2e-2$  \\
\hline
\hline
$w_k$ & $\phi_k$ & $\lambda_k$& $\phi_0$ \\
\hline
$0.99995578$ & $-0.00477923$ &$-0.40479907$& \\ 
$2.99892898$&$-3.139197051$ & $0.04915202$& $0.335023246$ \\ 
$3.99604127$& $0.01794715$ & $0.02874206$& \\ 
$ -1.99965386$& $0.00445702$ & $ 0.10497063$& \\ 
\hline
\end{tabular}
\vspace{3mm}
\caption{\;Number of iterations, value of the loss function at convergence, and optimal weights and biases of the FNN to approximate $ f(x) = x^2$ on $[-1, 1$].}\label{tab:tabfnnx2}
\end{center}
\end{table}

\begin{figure}
    \centering
    \includegraphics[width=.45\textwidth]{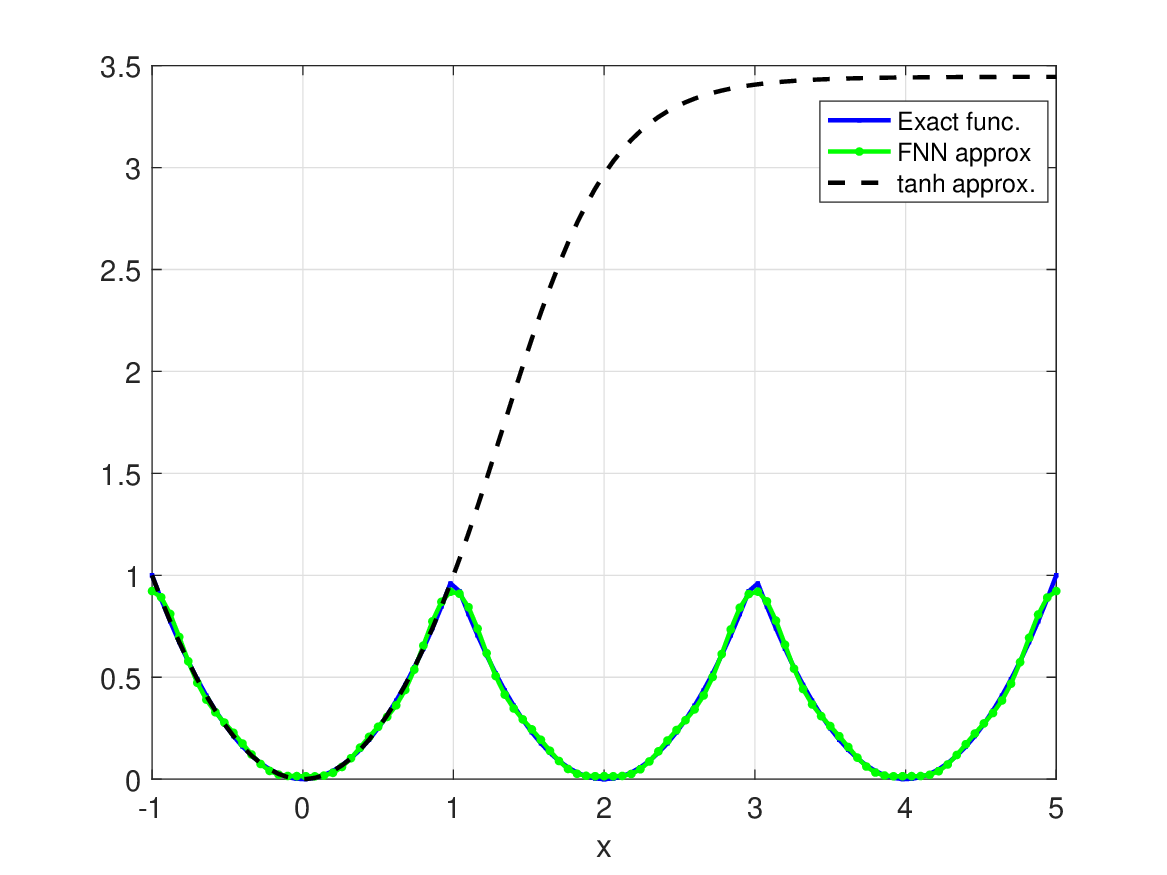}
    \caption{\;Comparison between the FNN and the $\mathrm{tanh}$ approximations outside the training domain for $f(x) = x^2$. }
    \label{fig:Fnnx2out}
\end{figure}

\begin{table}
  \begin{center}
  \begin{tabular}{ |c|c|c|c|c| } 
    \hline
\multicolumn{3}{|c|}{Number of iterations}& $445$  \\
\hline
  \multicolumn{3}{|c|}{Loss Function at convergence} & $1e-2$  \\
\hline
\hline
$w_k$ & $\phi_k$ & $\lambda_k$& $\phi_0$ \\
\hline
$ 0.99995402$ & $-2.44738021e-3$ &$-0.40420162$& \\ 
$2.98845687$&$-1.68031025e-1$ & $0.03295792$& $0.502531$ \\ 
$2.99445216$& $-6.78306499e-2$ & $-0.07711458$& \\ 
$ -1.99075052$& $-9.45244148$ & $ -0.00391551$& \\ 
\hline
\end{tabular}
\vspace{3mm}
\caption{\;Number of iterations, value of the loss function at convergence, and optimal weights and biases of the FNN to approximate $ f(x) = |x|$ on $[-1, 1$].}\label{tab:tabfnnabsx}
\end{center}
\end{table}

We illustrate in Figure\myref{fig:Fnnx2out} the behavior of the FNN output against the approximation from a neural network with a $\mathrm{tanh}$ activation function. In this case as well, the FNN network conserves the  properties of the function to be approximated outside of the training domain, while the $\mathrm{tanh}$ neural network does not. We note that the latter was more accurate in this case on the training interval $[-1, 1]$ ($6e-4$ with $3073$ iterations against $2e-2$ with $130$ iterations for the FNN).

We also assessed the function
\begin{equation}\label{eq:absx}
 g(x) = |x|, \;\; \text{$x \in \left(-(2k+1), (2k+1)\right)$}\;\;, k \in \mathbf{N} ,    
\end{equation}
which has the Fourier series
$$
S_N g(x)= \frac{1}{2} + \sum_{n=1}^{N} -4 \frac{1}{\pi^2 (2n - 1)} \mathrm{cos}(\pi (2n - 1)x).
$$

We summarize the FNN coefficients in Table\myref{tab:tabfnnabsx}. The FNN was able to capture the first $4$ modes of the Fourier decomposition and split the value of the $3^{rd}$ mode between two nodes in the hidden layer as showcased in Table\myref{tab:tabfnnabsx}. The reason is that $g$ admits Fourier coefficients only for odd coefficients. We provide in Figure\myref{fig:absx} a visual representation to compare the Fourier coefficients of $g$ with its FNN coefficients. Results similar to those obtained for the function $f$ defined in Equation\myref{eq:xsquare} are observed here.  The FNN coefficients are good approximations of their Fourier counterparts, and the error between the two has $1e-3$ order of magnitude.

\section{Fourier neural networks for differential equations solvers}\label{sec:fnnpde}
Machine learning in the field of differential equations fills various gap, from providing alternative models to the classical continuum mechanics equations, to inverse problem solvers, and thus has generated considerable interest in recent years. ML techniques have been used not only to effectively solve differential equations \cite{Dockhorn2019}, \cite{hsieh2019learning}, \cite{raissi2017hidden}, \cite{raissi2019Pinn},  and \cite{Sirignano2018} but also to discover underlying dynamics from data \cite{Brunton2016}, \cite{Raissi2018} and \cite{ Rudy2017}. A different approach is to build robust neural networks architectures based on differential equations \cite{Chen2018}, \cite{Long2018} and \cite{Ruthotto2019}. 

In this section, we extend the work presented in Section\myref{sec:fnn} to compute periodic solutions of differential equations of the type $Pu = f,$ where $P$ is a differential operator. To this end, we follow \cite{Raissi2018} and \cite{Sirignano2018} and set the loss function to be the residual of the differential equation we intend to solve. In a spirit similar to the weak formulations for finite element solutions \cite{Hutton2004}, the solution is obtained by minimizing $||P \hat{u}(x) - f(x) ||_2^2$. The concept was introduced in \cite{Sirignano2018} and was called physics informed neural networks in \cite{raissi2019Pinn}.  To leverage the Fourier neural networks introduced here, we append to the loss function the periodicity requirement as well as the regularization terms to obtain
\begin{equation}\label{eq:lossfunPDE}
     L(\phi, w, \lambda) = ||P \hat{u}(x) - f(x) ||_2^2  + \alpha_2||\lambda||_2^2 + \alpha_3||w||_2^2 + \alpha_3\left( ||\hat{u}(x + T) - \hat{u}(x)||_2^2\right) + \alpha_4\left( ||\hat{u}(x - T) - \hat{u}(x)||_2^2 \right).
\end{equation}
We refer to this residual-based loss function approach in conjunction with cosine as the activation function as physics informed Fourier neural network (PIFNN), to extend on previous work.

\subsection{Poisson equation}
  \begin{figure}
    \centering
     \subfloat[Comparison between the PIFNN and the exact solutions.]{\includegraphics[width=.45\textwidth]{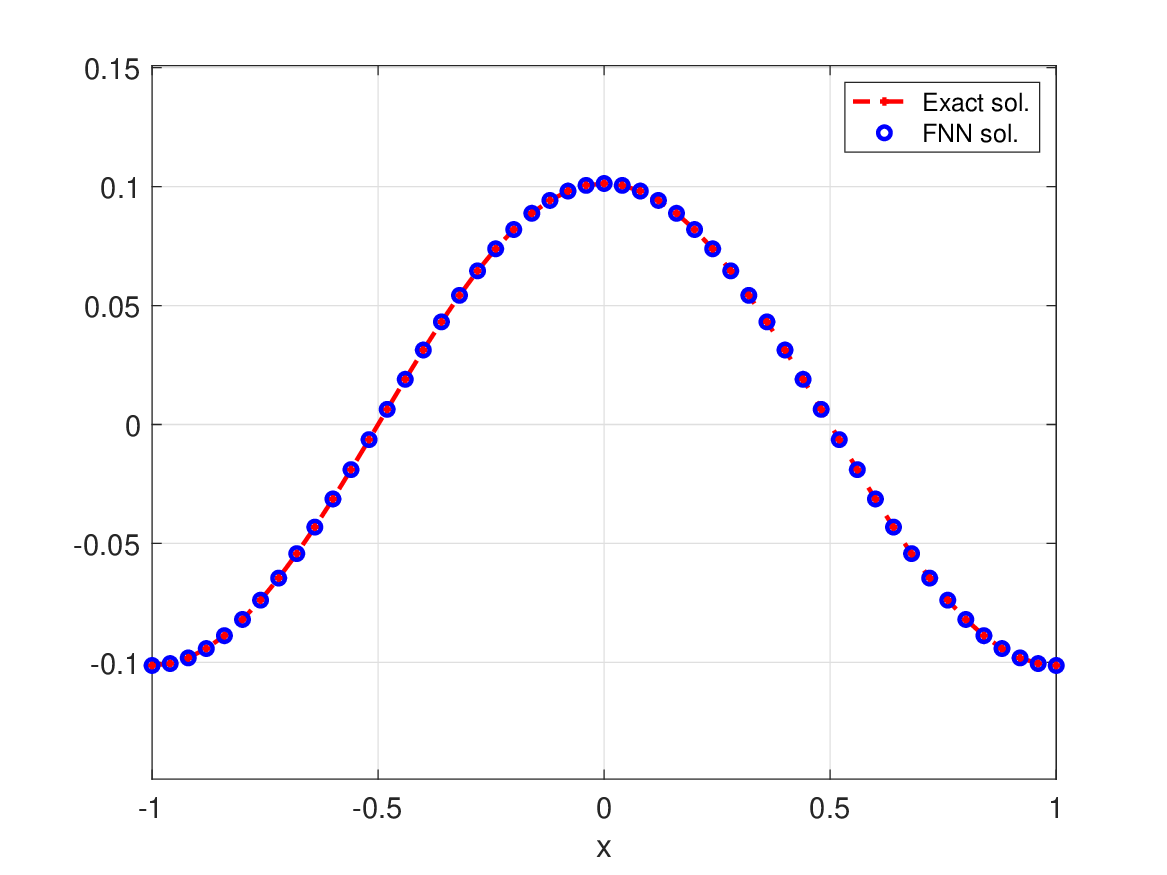}\label{fig:FNNvsexactPoisson_sol}}
     \subfloat[Pointwise relative error between the PIFNN and the exact solutions.]{\includegraphics[width=.45\textwidth]{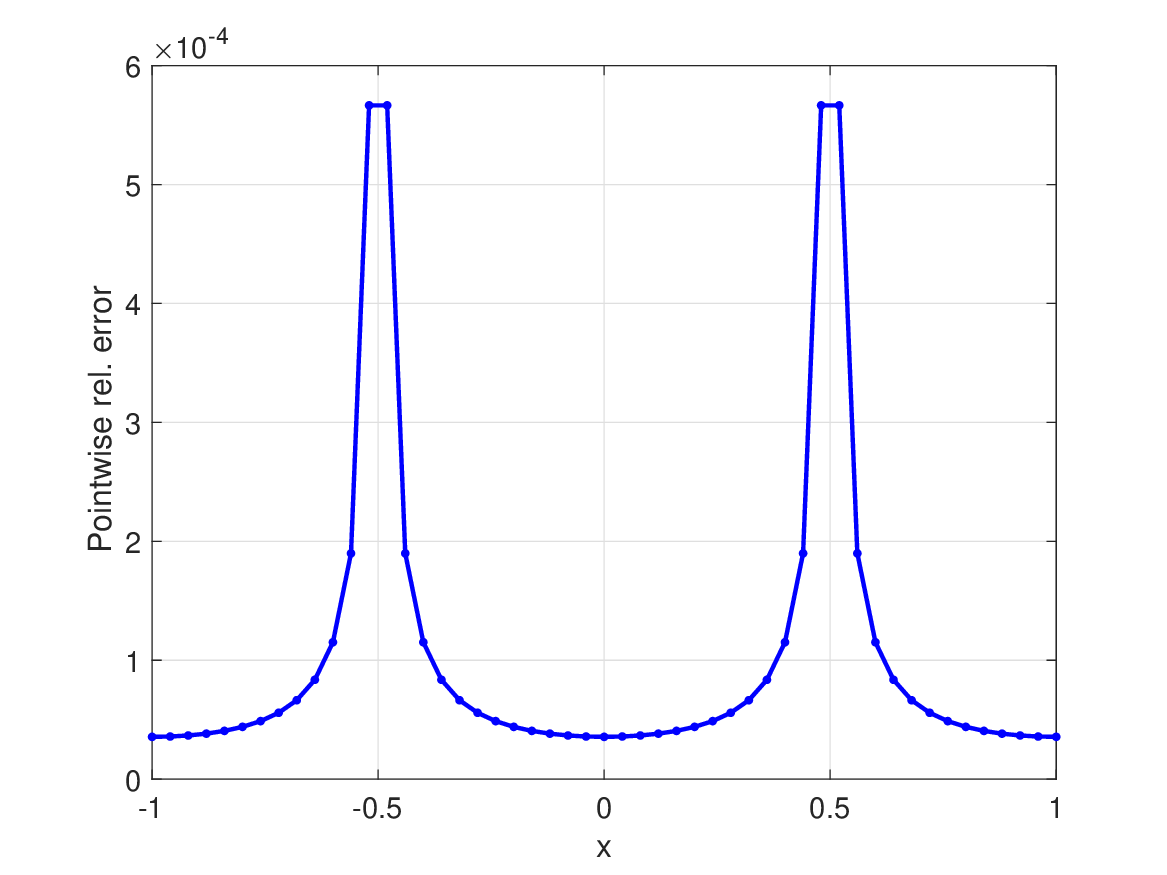}\label{fig:FNNvsexactPoisson_error}}
    \caption{\;Qualitative assessment of the PIFNN solution for the Poisson equation.}
    \label{fig:FNNvsexactPoisson}
\end{figure}

 \begin{table}
  \begin{center}
\begin{tabular}{ |c|c|c|c|c| } 
\hline
$w_k$ & $\phi_k$ & $\lambda_k$& $\phi_0$ \\
\hline
$-3.77424453e-10$ & $-4.42730681$ &$4.90842033e-5$& \\ 
$-1.16548318e-10$&$ 2.46210794$ & $-1.45660922e-5$& $0$ \\ 
$1.00000000$& $ 3.14159265 \approx \pi$ & $-1.01321184e-1 \approx -1/\pi^2$& \\ 
$ -1.18183192e-9$& $ 0.79153364 $ & $-1.60990031e-6$& \\ 
\hline
\end{tabular}
\vspace{3mm}
\caption{\;Optimal weights and biases of the PIFNN to solve the Poisson Equation\myref{eq:Poissoneq}, $k = 1\ldots4$. }\label{tab:tabPoisson}
\end{center}
\end{table}
Let us consider the Poisson equation with periodic boundary conditions,
\begin{equation}\label{eq:Poissoneq}
    -\Delta u = \mathrm{cos}(\pi x),\quad  x \in (-1,1), \quad \text{with } u(-1)=u(1),
\end{equation}
and compute its solution using the constructed PIFNN with four nodes in the hidden layer. The exact solution of the equation, given by $u = \frac{1}{\pi^2}\mathrm{cos}(\pi x)$, allows us to assess the accuracy of the PIFNN solution. We show both solutions in Figure\myref{fig:FNNvsexactPoisson_sol}, and we report the weights and biases of the PIFNN in Table\myref{tab:tabPoisson}. We note that the output is 
$$\hat{u}(x) = -\frac{1}{\pi^2}\mathrm{cos}(\pi x + \pi) + o(10^{-5}) \approx \frac{1}{\pi^2}\mathrm{cos}(\pi x).$$
and we observe that the pointwise relative error between the two solutions has order of magnitude $1e-4$ in Figure\myref{fig:FNNvsexactPoisson_error}. In contrast with the original PINN approach that provides an uninterpretable neural network model for the solution of the differential equation, PIFNN provides an analytic formula fully determined by the computed weights. 

\subsection{Heat equation}
To extend further the FNN framework to time-dependent problems, we consider the heat equation with periodic boundary conditions over an interval $[-1,\ 1]$. The solution is therefore $\tau=2$ periodic: 
\begin{align}\label{eq:Heateq}
    \frac{\partial u}{\partial t} = \frac{\partial^2 u}{\partial^2 x},\quad & x \in (-1,1), \; \;t \in [0,4] ,\nonumber \\
    u(0,x) = \mathrm{sin}(\pi x),\quad & u(t, -1) = u(t, 1). 
\end{align}

Various strategies are available for treating time-dependent PDEs by using neural networks \cite{raissi2019Pinn}, \cite{Sirignano2018}. Here we suggest an alternative idea to account for the time dependency, namely, to construct two separate networks for the spatial and time dimension. This can be achieved by rewriting the heat equation using the separation of variables method. To this end, we introduce two separate variables $X,T$ for the spatial and time component, respectively, and set $u(x,t) = X(x)T(t)$. Replacing the ansatz on $u$ in Equation\myref{eq:Heateq} transforms the equation into 
$$X(x)T'(t) = X''(x)T(t).$$

  \begin{figure}
    \centering
    \subfloat[PIFNN solution]{\includegraphics[width=0.45\linewidth]{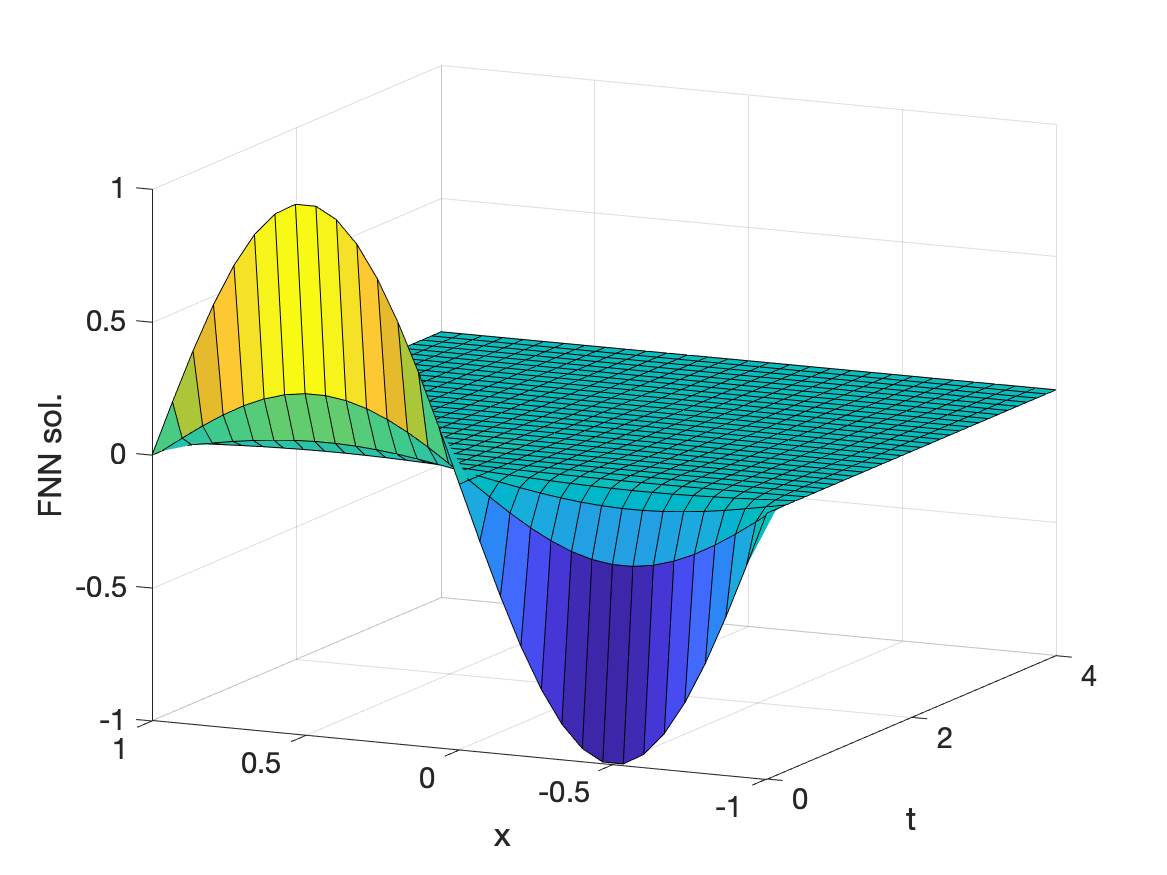}\label{fig:fnnvsexactheat_sol}}
     \subfloat[Pointwise error between the PIFNN and the exact solutions]{\includegraphics[width=0.45\linewidth]{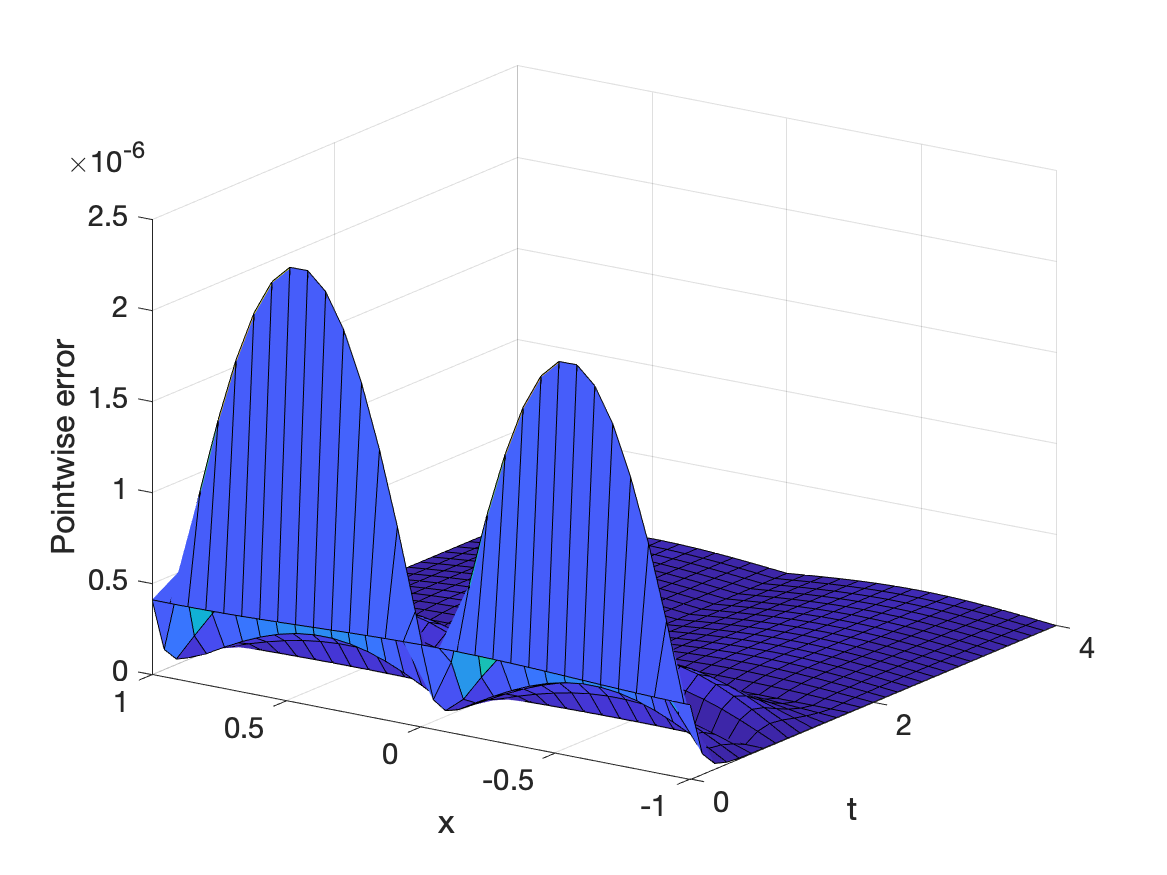}\label{fig:fnnvsexactheat_err}}
     \hfill
     \subfloat[Pointwise error between the PDEPE and the exact solutions]{\includegraphics[width=0.45\linewidth]{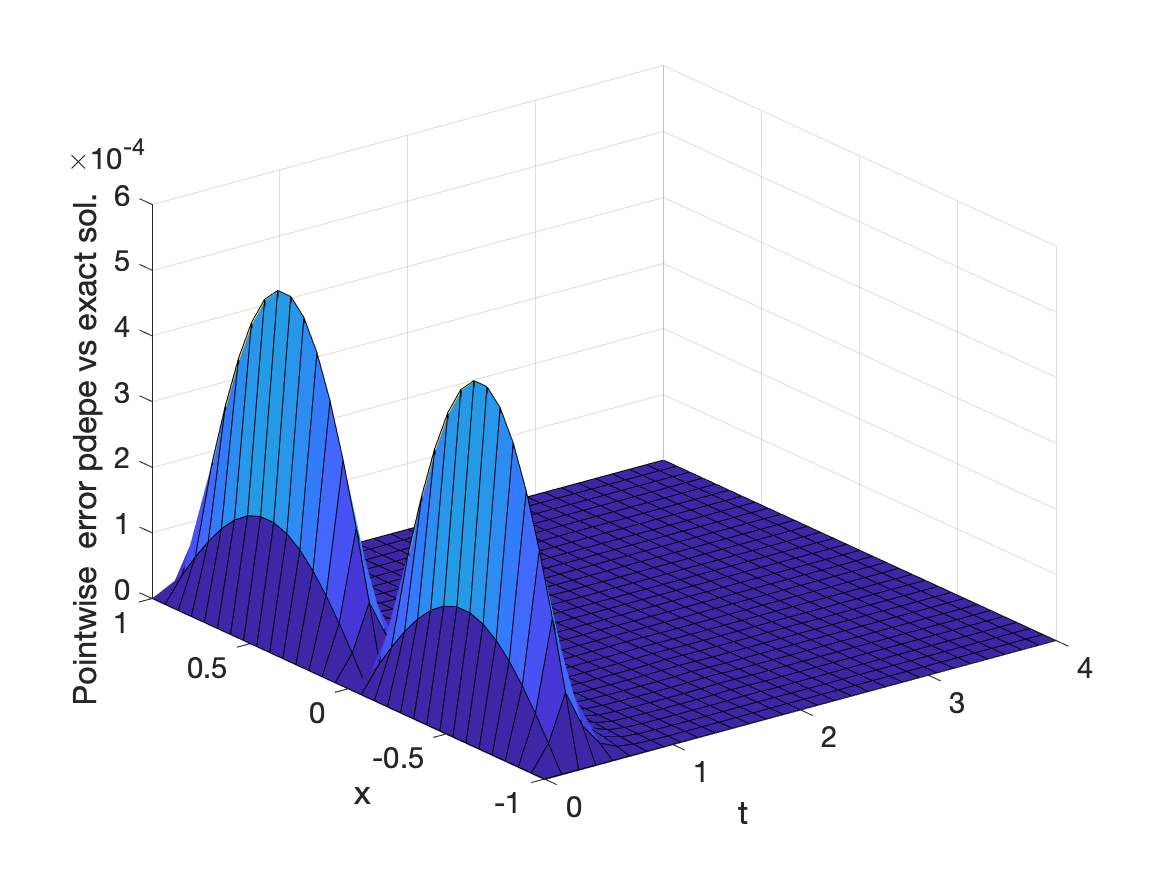}\label{fig:pdepevsexactheat_err}}
     \subfloat[Pointwise error between the PIFNN and the PDEPE solutions]{\includegraphics[width=0.45\linewidth]{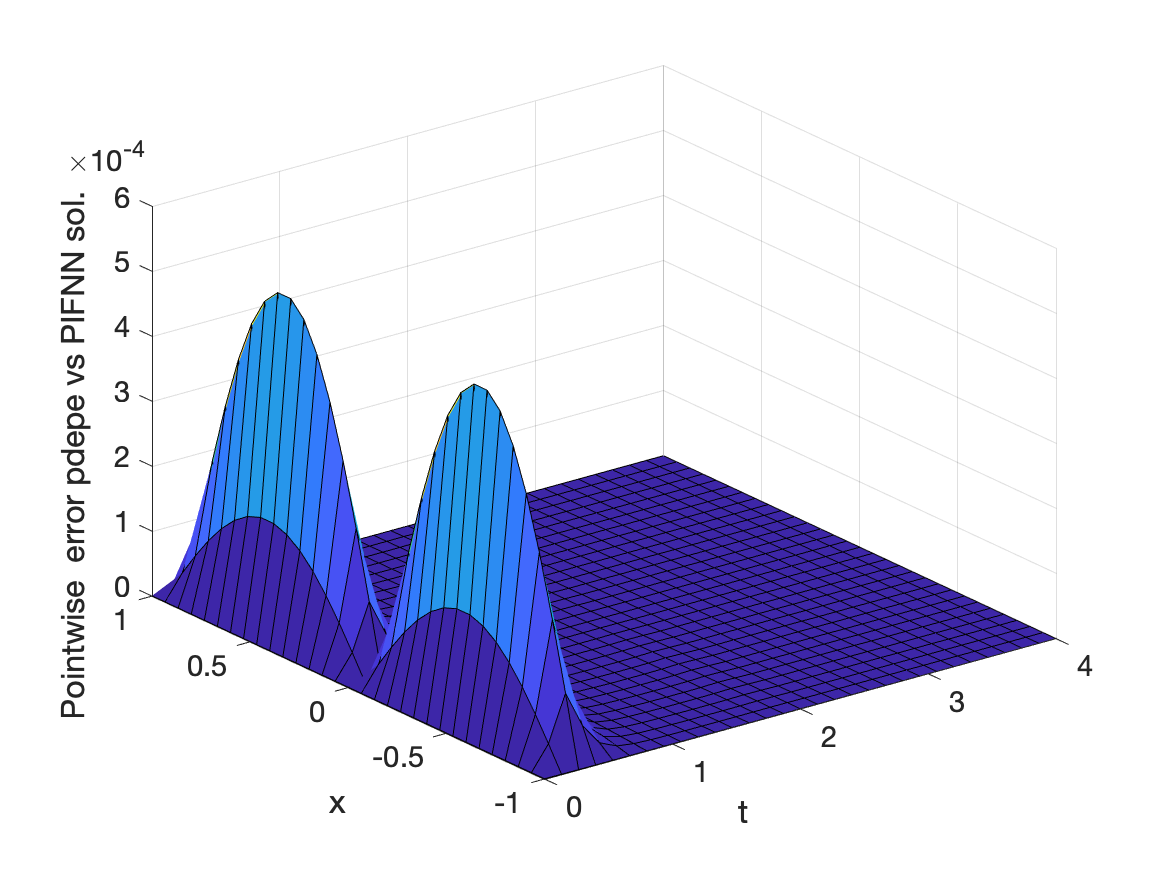}\label{fig:pifnnvspdepeheat_err}}
    \caption{\;Qualitative assessment of the PIFNN solution for the heat equation.}
    \label{fig:fnnvsexactheat}
\end{figure}
Note that this strategy is not universally applicable, but it is a common approach used to derive analytical solutions to linear differential equations.
We rewrite the cost function as 
\begin{align*}
    L(\phi, w, \lambda) = &\alpha_1||\hat{X}(x)\hat{T}'(t)-\hat{X}''(x)\hat{T}(t)||^2 +\alpha_2||\lambda||^2 + \alpha_3||w||^2 \\ 
    &+\alpha_3 ||\hat{X}(x + 2) - \hat{X}(x)||^2 +\alpha_4||\hat{X}(x - 2) - \hat{X}(x)||^2  + \alpha_5||\hat{T}_0(x) - \hat{u}_0(x)||^2 ,
\end{align*}
to incorporate the boundary and initial conditions and separate the network into two separate subnetworks. The spatial components are modeled by using the PIFNN approach with one node in the hidden layer, while the time component is treated by using a shallow neural network with a $\mathrm{tanh}$ activation function and five nodes in the hidden layer. 

Figure\myref{fig:fnnvsexactheat} provides a qualitative assessmeent of the PIFNN solution for the heat Equation\myref{eq:Heateq}.  Figure\myref{fig:fnnvsexactheat_sol} corresponds to the PIFNN solution and Figure\myref{fig:fnnvsexactheat_err} the error between that solution and the exact one. We appended a comparison with the solution to Equation\myref{eq:Heateq} provided by the Matlab \textit{pdepe} solver \cite{MATLAB:2018}. The error between the PIFNN and the exact solution is of magnitude $1e-6$ and the one between the \textit{pdepe} solution and the exact one is of magnitude $1e-4$. The model chosen here is likely to ensure smoothness given that errors in Fourier coefficients are mollified by the activation function.

\section{Conclusion}
\label{sec:conc}
We have constructed a novel Fourier neural network model that successfully approximates low-frequency analytic and piecewise continuous one-dimensional periodic functions.  By combining the FNN with a residual-based loss function, as in the PINN model, we define a fully interpretable network for differential equations with periodic boundary conditions, which we refer to as PIFNN. The FNN model replaces commonly used activation functions, such as $\mathrm{tanh}$ or sigmoid, by a periodic function. With cosine as the activation function, the network model mimics a Fourier decomposition. To ensure the periodic nature is preserved, we appended to the loss function the requirement that the solution be periodically replicated also outside the training domain. 
Besides significantly improving the results obtained using traditional neural networks in terms of accuracy and number of iterations, our simulations revealed another significant advantage: namely, FNN conserves the properties of the learned task (approximating a function, etc.) outside the training domain.

One limitation of this model is that we do not use a rigorous technique for picking the penalty coefficients but set them heuristically. Another limitation is that we need to know the period of the function we estimate a priori to incorporate it in the activation function and loss function. We are currently investigating solutions to overcome these limitations. Our results are promising, and we plan on extending them to multidimensional functions and nonlinear differential equations. We believe the framework presented here can serve as a pre- and postprocessing tool in many engineering and scientific applications, where the Fourier decomposition is widely used, such as electronics, image recognition, and acoustics.

\section*{Acknowledgments}
%GAIL - note that the govt license shold go at th end of the document and is removed upon publication
 We thank Charlotte Haley for her pertinent suggestions. This material is based upon work supported by the
U.S. Department of Energy, Office of Science, Advanced Scientific
Computing Research under Contract DE-AC02-06CH11357. \\
{\bf Government License.}  The submitted manuscript has been created by
UChicago Argonne, LLC, Operator of Argonne National Laboratory
(``Argonne''). Argonne, a U.S. Department of Energy Office of Science
laboratory, is operated under Contract No. DE-AC02-06CH11357. The
U.S. Government retains for itself, and others acting on its behalf, a
paid-up nonexclusive, irrevocable worldwide license in said article to
reproduce, prepare derivative works, distribute copies to the public,
and perform publicly and display publicly, by or on behalf of the
Government.  The Department of Energy will provide public access to
these results of federally sponsored research in accordance with the
DOE Public Access
Plan. http://energy.gov/downloads/doe-public-access-plan.

\appendix
%\renewcommand\appendixname{Appendix}
%\section{The first appendix} 

%\begin{appendices}
\section{Statistical properties}\label{appendixA}
We recall below some properties verified by the mean $\mu$ and the variance $\sigma^2$ of a random variable.
\begin{enumerate}%[label=Property \arabic*.,itemindent=*]
  \item $\mu(aX + Y) = a\mu(X) + \mu(Y)$ and $\sigma^2(aX+Y) = a^2\sigma^2(X) + \sigma^2(Y)$, where $a \in \mathbf{R}$ and $X$ and $Y$ are two random variables.
  \item $\mu(XY) = \mu(X) \mu(Y)$ and $\sigma^2(XY) = \sigma^2(X) \left(\mu^2(Y) + \sigma^2(Y)\right) + \mu^2(X)\sigma^2(Y)$, where $X$ and $Y$ are two independent random variables.
\end{enumerate}
We also recall the value of the mean and of the variance of the two distributions used in the paper, namely, the uniform and normal distributions. 
\begin{enumerate}
  \item If $X \sim \mathcal{U}([a,b])$ then:
  \begin{itemize}
      \item $\mu(X) = \frac{1}{b - a}$ and  $\sigma^2(X) = \frac{(b - a)^2}{12}$.
      \item The  probability density function (pdf) of $X$ is $f_X(x) = \frac{1}{b-a}$ if $x \in [a,b]$ and $0$ otherwise.
  \end{itemize}
   
  \item If $X \sim \mathcal N (0, m^2)$, then
   \begin{itemize}
      \item $\mu(X) = 0$ and  $\sigma^2(X) = m^2$, 
      \item The  probability density function (pdf) of $X$ is $f_X(x) = \frac{1}{\sigma \sqrt{2\pi}}e^{-\frac{x^2}{2\sigma^2}}$.
  \end{itemize}
\end{enumerate}
%\end{appendices}

\bibliographystyle{plain}
\bibliography{Ngom_Marin_FNN_revisions_v2}

\end{document}